\documentclass[final,3p,times,twocolumn,authoryear]{elsarticle}
\usepackage{algorithmic}
\usepackage{algorithm}
\usepackage{array}
\usepackage[caption=false,font=normalsize,labelfont=sf,textfont=sf]{subfig}
\usepackage{textcomp}
\usepackage{stfloats}
\usepackage{url}
\usepackage{verbatim}
\usepackage{graphicx}
\usepackage{bm}
\usepackage{makecell}

\usepackage{algorithm}
\usepackage{algorithmic}
\usepackage{siunitx}   % 用于书写 数字+单位 的格式
\usepackage{upgreek}   % 用于书写 数字+单位 的格式
\usepackage{xcolor}    % colors
\usepackage{booktabs}       % professional-quality tables
\usepackage{longtable}
\usepackage{threeparttable}
\usepackage{multirow}
\usepackage{multicol}
\usepackage{times}
\usepackage{hyperref}
\usepackage{enumitem}

\usepackage{natbib}
\usepackage{amsthm}
\newtheorem{theorem}{Theorem}
\newtheorem{lemma}{Lemma}

\def\N{\mathbf{N}}

\def\N{\mathbf{N}}

\def\N{\mathbf{N}}

\def\N{\mathbf{N}}

\def\N{\mathcal N}

\def\0{\mathbf{0}}

\definecolor{darkblue}{rgb}{0,0.22,0.66}
\hypersetup{colorlinks=true,citecolor=cyan,linkcolor=darkblue}
\usepackage{mathrsfs}
\usepackage{float}%% 
\usepackage{mathtools,amssymb}
\graphicspath{{figuresPF/}{figure/}}
\let\cdots\customcdots
\let\eps\varepsilon
\def\tildeeps{{\widetilde{\eps}}}
\let\tn\textnormal
\let\myforall\forall
\def\forall{{\myforall\, }}
\let\myexists\exists
\def\exists{{\myexists\, }}

\def\bmx{{\bm{x}}}
\def\bmphi{{\bm{\phi}}}
\newcommand{\bmzero}{{\bm{0}}}
\newcommand{\bmlambda}{{\bm{\lambda}}}
\def\bbR{{\mathbb{R}}}
\def\bbQ{{\mathbb{Q}}}
\def\bbN{{\mathbb{N}}}

\def\calO{{\mathcal{O}}}
\def\calI{{\mathcal{I}}}
\def\calL{{\mathcal{L}}}
\def\scrA{{\mathscr{A}}}
\def\caltildeI{{\widetilde{\mathcal{I}}}}

\let\barscrA\scrbarA
\def\tildephi{{\widetilde{\phi}}}
\def\tildevarrho{{\widetilde{\varrho}}}
\def\tildeN{{\widetilde{N}}}
\def\tildeL{{\widetilde{L}}}
\newcommand{\dprime}{{\prime\prime}}

\usepackage[most]{tcolorbox}
\definecolor{lightblue}{RGB}{225, 246, 252}
\definecolor{lightyellow}{RGB}{255, 255, 230}

\newtcolorbox{tcboxThm}{
	center,
	breakable,
	boxrule=0.6pt,
     colback=blue!6, 
     colframe=white,
     arc=2mm, auto outer arc,
     left=5pt,
    right=5pt,
    top=4pt,
    bottom=4pt,
}

\newtcolorbox{tcboxLem}{
	center,
	breakable,
	boxrule=0.6pt,
     colback=black!6, 
     colframe=white,
     arc=2mm, auto outer arc,
          left=5pt,
    right=5pt,
    top=4pt,
    bottom=4pt,
}

\newcommand{\nnOneD}[6][]{\ensuremath{
{\hspace{0.6pt}\mathcal{N}\hspace{-1.98pt}\mathcal{N}\hspace{-0.725pt}}_{#2}%  NN
		#1\{#3,\hspace{1.987pt} #4;\hspace{3.0297pt} {#5}\hspace{-1.0298pt}\to\hspace{-0.98pt}{#6}#1\}
}}

\usepackage{amssymb}
\journal{Neural Networks}

\begin{document}
% \hbadness=10000
\begin{frontmatter}



\title{ Don't Fear Peculiar Activation Functions: EUAF and Beyond}


\author[inst1]{Qianchao Wang\corref{cor1}} %% Author name
\affiliation[inst1]{organization={Center of Mathematical Artificial Intelligence},%Department and Organization
            addressline={Department of Mathematics}, 
            city={The Chinese University of Hong Kong},
            % postcode={999077}, 
            state={Hong Kong},
            country={China}}
% \ead{230208053@seu.edu.cn}
\author[inst2]{Shijun Zhang\corref{cor1}} %% Author name
\affiliation[inst2]{%Department and Organization
            addressline={Department of Applied Mathematics}, 
            city={The Hong Kong Polytechnic University},
            % postcode={999077}, 
            state={Hong Kong},
            country={China}}
\author[inst3]{Dong Zeng} %% Author name
\affiliation[inst3]{%Department and Organization
            addressline={Department of Biomedical Engineering}, 
            city={Southern Medical University},
            % postcode={999077}, 
            state={Guangzhou},
            country={China}}
\author[inst4]{Zhaoheng Xie} %% Author name
\affiliation[inst4]{organization={Institute of Medical Technology},%Department and Organization
            addressline={Peking University Health Science Center}, 
            city={Peking University},
            % postcode={999077}, 
            state={Beijing},
            country={China}}
\author[inst5]{Hengtao Guo} %% Author name
\affiliation[inst5]{organization={Independent Researcher},%Department and Organization
            addressline={708 6th Ave N, Seattle, WA 98109, US}} 
%             % city={Peking University},
%             % postcode={999077}, 
%             % state={Beijing},
%             country={US}}         
\author[inst1]{Tieyong Zeng}
\author[inst1]{Feng-Lei Fan\corref{mycorrespondingauthor}} %% Author name
\cortext[mycorrespondingauthor]{Corresponding author, hitfanfenglei@gmail.com}
 %% Author name

\cortext[cor1]{Qianchao Wang and Shijun Zhang are co-first authors.}

% \author{Qianchao Wang$^{1*}$, Shijun Zhang$^{2*}$, Dong Zeng$^{3}$, Zhaoheng Xie$^{4}$, Hengtao Guo$^{5}$, Feng-Lei Fan$^{1\dag}$, \textit{Member, IEEE}, Tieyong Zeng$^{1}$ % <-this % stops a space
% \thanks{*Qianchao Wang and Shijun Zhang are co-first authors. Feng-Lei Fan is the corresponding author (hitfanfenglei@gmail.com) }% <-this % stops a space
% \thanks{$^{1}$Qianchao Wang, Tieyong Zeng, and Feng-Lei Fan are with Center of Mathematical Artificial Intelligence, Department of Mathematics, The Chinese University of Hong Kong, Shatin, Hong Kong.}
% \thanks{$^{2}$Shijun Zhang is with Department of Mathematics, Duke University, Duram, NC, US} 
% \thanks{$^{3}$Dong Zeng is with Department of Biomedical Engineering, Southern Medical University, Guangzhou, China} 
% \thanks{$^{4}$Zhaoheng Xie is with Institute of Medical Technology, Peking University Health Science Center, Peking University, Beijing, China.}
% \thanks{$^{5}$Hengtao Guo is an independent researcher.}
% }

%% Abstract
\begin{abstract}

In this paper, we propose a new super-expressive activation function called the Parametric Elementary Universal Activation Function (PEUAF). We demonstrate the effectiveness of PEUAF through systematic and comprehensive experiments on various industrial and image datasets, including CIFAR10, Tiny-ImageNet, and ImageNet. Moreover, we significantly generalize the family of super-expressive activation functions, whose existence has been demonstrated in several recent works by showing that any continuous function can be approximated to any desired accuracy by a fixed-size network with a specific super-expressive activation function. Specifically,  our work addresses two major bottlenecks in impeding the development of super-expressive activation functions: the limited identification of super-expressive functions, which raises doubts about their broad applicability, and their often peculiar forms, which lead to skepticism regarding their scalability and practicality in real-world applications.

% Despite the theoretical elegance of super-expressive activation functions, their development faces two major bottlenecks: Firstly, only a limited number of super-expressive functions have been identified so far. This raises the question of whether the super-expressive property can be broadly applied. Secondly, the super-expressive activation function usually has a peculiar form, which makes most practitioners skeptical about its scalability and practicality in real-world applications. In this paper, to confront these two bottlenecks, we first substantially generalize a kind of activation function to encompass a large family of functions capable of achieving super-expressiveness. Then, we introduce a new super-expressive activation function, referred to as the Parametric Elementary Universal Activation Function (\texttt{PEUAF}), and validate the effectiveness of \texttt{PEUAF} using systematic and comprehensive experiments on four industrial datasets and three image datasets (CIFAR10, Tiny-ImageNet, and ImageNet). Future research could focus on further extending the family of super-expressive activation functions and elaborating their practical utility in more diverse and complex datasets.

\end{abstract}


% \begin{highlights}

% \item We bridge the gap between the theoretical elegance and empirical usefulness of super-expressive functions.  

% \item We provide a non-trivial generalization of EUAF, showing that a broader family of activation functions can achieve super-expressiveness.
 
% \item We demonstrate the superiority of PEUAF through systematic experiments on four industrial datasets and three image datasets. 

% \end{highlights}
%%%%%%%%%%%%%
%% Keywords

\end{frontmatter}



\section{INTRODUCTION}

% In recent years, deep learning has achieved great success across numerous critical domains~\citep{lecun2015deep}. A key factor contributing to the success of deep learning is the development of highly effective nonlinear activation functions, which significantly improve the neural network's information processing capabilities. Despite well-established options like the Rectified Linear Unit (\texttt{ReLU}) and its variants ~\citep{nair2010rectified}, given the fundamental importance of activation functions, searching for optimal ones is an endless endeavor, and researchers are constantly striving to design and evaluate various activation functions, both through theoretical analysis and empirical studies ~\citep{bingham2022discovering,apicella2021survey,wang2024dynamics}.

In recent years, deep learning has achieved significant success in many critical areas~\citep{lecun2015deep}. A major factor contributing to this success is the development of highly effective nonlinear activation functions, which greatly enhance the information processing capabilities of neural networks. While established options like the Rectified Linear Unit (\texttt{ReLU}) and its variants are widely used~\citep{nair2010rectified}, the fundamental importance of activation functions makes the search for better ones a continuous effort. Researchers are persistently working to design and evaluate various activation functions through both theoretical analysis and empirical studies~\citep{bingham2022discovering,apicella2021survey,wang2024dynamics}.

% In this context, the approximation theory has shown that there exists a special kind of activation function that allows a network with simple structures to approximate any continuous function with an arbitrarily small error using a fixed number of neurons \citep{MAIOROV199981}. These activation functions are known as ``super-expressive activation functions'' ~\citep{yarotsky2021elementary}. Research indicates that an activation function must have both the stationary part and analytical part to achieve the super-expressiveness~\citep{2021Deep,yarotsky2021elementary}, \textit{i.e.}, the elementary universal activation function (\texttt{EUAF}). As shown in Figure~\ref{fig:EUAF}, the expression of \texttt{EUAF} is 
% \begin{equation*}%\label{eq:def:sigma}
% 	\mathrm{EUAF}(x)\coloneqq \begin{cases}
% 		\big|x-2\lfloor \tfrac{x+1}{2}\rfloor\big|  & \tn{for} \  x\ge  0,\\
% 		\frac{x}{1+|x|} & \tn{for} \  x<0,\\
% 	\end{cases}
% \end{equation*}
% where the values to the left of zero are analytical and those to the right are stationary. The property of super-expressiveness is not only unique but also highly desirable. To the best of our knowledge, in most universal approximation constructs, the network structures and the number of neurons have to become more complex as the approximation error decreases. In contrast, this limitation can be surprisingly overcome by simply replacing the activation functions with super-expressive ones. This means that adjusting parameters can achieve any pre-specified approximation accuracy without increasing the network complexity. 

\begin{figure}[htbp!]        
	\centering          
	\includegraphics[width=0.8\linewidth]{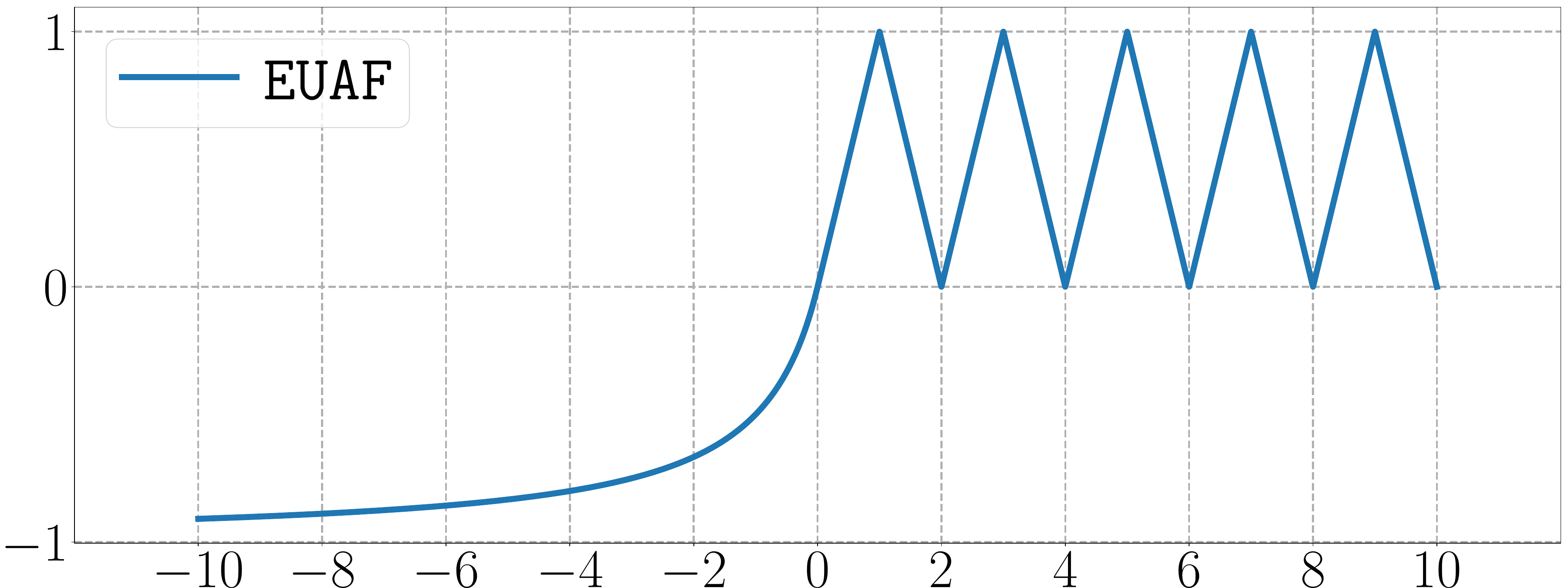}
	\caption{An illustration of $\mathrm{EUAF}$.}
	\label{fig:EUAF}
 \vspace{-0.3cm}
\end{figure}

In the realm of approximation theory, it has been shown that certain activation functions can empower a neural network with a simple structure to approximate any continuous function with an arbitrarily small error, using a fixed number of neurons~\citep{MAIOROV199981}. These functions are termed ``super-expressive activation functions"~\citep{yarotsky2021elementary}. According to research, to achieve super-expressiveness, an activation function should possess both periodic and analytical components~\citep{2021Deep,yarotsky2021elementary}. One such example is the elementary universal activation function (\texttt{EUAF}), defined as follows:
\begin{equation*}
	\mathrm{EUAF}(x)\coloneqq \begin{cases}
		\big|x-2\lfloor \tfrac{x+1}{2}\rfloor\big|  & \text{for} \  x\ge  0,\\
		\frac{x}{1+|x|} & \text{for} \  x<0,\\
	\end{cases}
\end{equation*}
Figure~\ref{fig:EUAF} depicts \texttt{EUAF}, an analytical function on $(-\infty,0)$ and periodic on $[0,\infty)$.
% .
% Figure~\ref{fig:EUAF} depicts the \texttt{EUAF},
% In this function, values for \( x < 0 \) are analytical, while values for \( x \geq 0 \) are stationary.
The unique and highly desirable property of super-expressiveness allows neural networks to achieve precise approximation accuracy without increasing network complexity. This contrasts with traditional universal approximation methods, where more complex structures and a higher number of neurons are required as the approximation error decreases. By integrating super-expressive activation functions, one can attain the desired approximation accuracy by merely adjusting parameters, thus maintaining a simpler network architecture.

To the best of our knowledge, the development of super-expressive activation functions faces two technical challenges that hinder their potential value to neural networks: 1) First, only a limited number of super-expressive functions have been identified so far ~\citep{MAIOROV199981,2021Deep,yarotsky2021elementary}. It is unclear if the super-expressive property can be broadly applied. Additionally, for deep learning practitioners, having a greater variety of activation functions that exhibit learning capabilities is necessary in terms of enriching their armory. Developing more super-expressive functions increases the likelihood of finding their utilities in important applications, as different activation functions differ in their trainability. 2) Second, the practical utility of super-expressive activation functions is questionable. While superior expressiveness can be theoretically established through specialized constructions that demonstrate the existence of an expressive solution ~\citep{Shen_2021,yarotsky2021elementary}, this does not necessarily translate to better practical performance. Furthermore, it is unclear whether gradient-based methods can effectively learn good solutions for networks using these functions.

Compared to commonly used functions like \texttt{ReLU}, \texttt{sigmoid}, and \texttt{tanh}, super-expressive functions usually have peculiar shapes. For example, Figure~\ref{fig:EUAF} shows \texttt{EUAF}, which is a typical super-expressive activation function. It has a complex and intimidating form, which makes most practitioners skeptical about its scalability and practicality in real-world applications. If we can demonstrate the practical utility of any super-expressive activation function, it could help resolve the skepticism and bridge the gap between their theoretical elegance and usefulness.

In addressing the first bottleneck, we substantially generalize the scope of \texttt{EUAF} to encompass a large family of functions capable of achieving super-expressiveness. Specifically, an activation function $\rho$ is considered to be super-expressive if it is real analytic within a small interval and a fixed-size $\rho$-activated network can reproduce a triangle-wave function. To address the second bottleneck, we believe that super-expressive functions can indeed be practically useful. Previous studies ~\citep{sitzmann2020implicit, ramirez2023deep} successfully applied the periodic function \texttt{sin} as an activation function within the implicit neural representation. These models have been demonstrated to be suitable for representing complex signals and their derivatives, as well as for solving challenging boundary value problems ~\citep{liu2022multistability}. These studies provide valuable insights into the potential of super-expressive activation functions, since both super-expressive activation functions and \texttt{sin} share periodicity. Moreover, from the perspective of signal decomposition, normal activation functions like \texttt{ReLU} tend to assist models in identifying the direct component (DC) of a signal ~\citep{lee2024magnitude}. In contrast, super-expressive activation functions can better handle stationary signals due to their inherent periodicity. This characteristic enhances their ability to manage complex real-world signals more efficiently.

Specifically, we choose \texttt{EUAF} as our representative and investigate a parameterized variant, named \texttt{PEUAF}, which adaptively learns the frequency $w$ on the positive side. Mathematically,
\begin{equation*}%\label{eq:def:sigma}
	\mathrm{PEUAF}(x)\coloneqq \begin{cases}
		\big|wx-2\lfloor \tfrac{wx+1}{2}\rfloor\big|  & \tn{for} \  x\ge  0,\\
		\frac{x}{1+|x|} & \tn{for} \  x<0,\\
	\end{cases}
\end{equation*}
where $w$ is the trainable parameter representing the frequency on the positive side. \texttt{PEUAF} can adaptively extract the stationary signals with different frequencies. This adaptability allows \texttt{PEUAF} to effectively capture and represent signals with diverse frequency components, which is particularly advantageous in addressing real-world signal complexities. Then, we validate the effectiveness of \texttt{PEUAF} by experimenting with four industrial datasets (1D data) and three image datasets (2D data). 
% For industrial datasets, \texttt{PEUAF} outperforms most state-of-the-art activation functions in terms of test accuracy, convergence speed, and the ability to localize faults. 
For industrial datasets, our tests show that \texttt{PEUAF} surpasses other activation functions in terms of test accuracy, convergence speed, and fault localization ability.
For image datasets, we find that combining \texttt{PEUAF} with other activation functions can usually yield better performance than only using a single activation function, although using \texttt{PEUAF} alone cannot achieve satisfactory performance. Thus, \texttt{PEUAF} can serve as a valuable add-on to the network. Our main contributions are as follows:

\begin{itemize}

\item We provide a non-trivial generalization of \texttt{EUAF}, showing that a broader family of activation functions can achieve super-expressiveness. 

\item We bridge the gap between the theoretical elegance and empirical usefulness of super-expressive functions by demonstrating their competitive performance in practical applications through systematic experiments on four industrial datasets and three image datasets including ImageNet.  
 
\item We introduce \texttt{PEUAF}, a parameterized version of \texttt{EUAF}, and demonstrate that \texttt{PEUAF} can be used individually or in conjunction with other well-performing activation functions.

\end{itemize}

\section{RELATED WORK}
\label{sec:related_work}
% Activation function is one of the key factor in neural network approximation, giving them their extraordinary ability to model complex functions and handle complex patterns~\citep{szandala2021review}. They serve as the nonlinear components in neural networks, elevating the neural networks beyond linear models~\citep{dubey2022activation}. In this section, we simply reviewed the conventional activation function  \texttt{ReLU} and its variant in Section~\ref{ReLU}. And then the rest activation functions are divided into tailored activation functions and super-expressive activation functions concluded in Section~\ref{Other AFs} and Section~\ref{SuperExp AFs}.

In the field of artificial intelligence, deep neural networks have proven to be highly effective tools. These networks leverage the power of interconnected nodes structured in multiple layers, allowing them to excel in a wide range of complex applications and new domains. At their core, deep neural networks rely on an affine linear transformation followed by a nonlinear activation function. The nonlinear activation function is essential for the successful training of these networks. 

% In recent years, the Rectified Linear Unit (\texttt{ReLU}) has become increasingly popular, demonstrating its effectiveness as an activation function.

Later in this section, we will first review conventional activation functions including \texttt{ReLU} and its variants, as well as recent sigmoidal activation functions in Section~\ref{ReLU}. We will then discuss super-expressive activation functions in Section~\ref{SuperExp AFs}.

\subsection{Conventional Activation Functions} \label{ReLU}

% In recent years, the Rectified Linear Unit (\texttt{ReLU}) has become increasingly popular, demonstrating its effectiveness as an activation function.
% $\mathrm{ReLU}(x)=\max(0,x)$ has been recognized as an efficient activation function by successfully addressing issues of gradient vanishment and explosion in \texttt{Sigmoid} and \texttt{Tanh} ~\citep{nair2010rectified}.
In recent years, the Rectified Linear Unit (\texttt{ReLU}~\citep{nair2010rectified}), defined as $\texttt{ReLU}(x)=\max(0,x)$, has gained popularity and recognition for its effectiveness in addressing the gradient vanishing and explosion issues encountered with \texttt{Sigmoid} and \texttt{Tanh} activation functions.
Thus, \texttt{ReLU} has been widely used in the deep learning community such as industrial fault diagnosis \citep{liu2024grid} and medical image segmentation \citep{liu2024segmenting} . However, \texttt{ReLU} can suffer from the occurrence of a number of ``dead neurons'', which results in information loss and can hurt the neural network's feature processing ability. To mitigate this issue, several variants of \texttt{ReLU} have been introduced such as Leaky Rectified Linear Unit (\texttt{LReLU}) ~\citep{2015Empirical}, Parametric Rectified Linear Unit (\texttt{PReLU}) ~\citep{he2015delving}, Randomized Leaky Rectified Linear Unit (\texttt{RReLU}) ~\citep{2015Empirical}, Exponential Linear Unit (\texttt{ELU}) ~\citep{clevert2015fast}, Gaussian Error Linear Unit (\texttt{GELU}) ~\citep{2016arXiv160608415H}, and Generalized Linear Unit (\texttt{GENLU}) ~\citep{fan2020soft}. Most recently, \cite{10572228} proposed Self-Normalizing \texttt{ReLU} or \texttt{NeLU} to ensure that the prediction model is not affected by the noise level during testing. It has been tested in synthetic data and image de-noising tasks. These variants represents a significant advancement in activation function design, offering adaptability and potentially better performance. Whereas, their benefits come with the cost of increased model complexity or computation burden and the need for careful tuning and regularization which inspired researchers to create more different activation functions.

In addition to these \texttt{ReLU} variants, other kinds of activation functions have also been developed. For example, the \texttt{Swish} ($\mathrm{Swish}(x)=x\cdot\mathrm{sigmoid}(\beta x)$) ~\citep{ramachandran2017searching} was identified through an automated search using a combination of exhaustive and reinforcement learning as an alternative to \texttt{ReLU}. Its similar shape makes it a reasonable proxy for \texttt{ReLU} in deep learning applications. \texttt{Mish}, defined as $\mathrm{Mish}(x)=x\cdot\mathrm{tanh}(\mathrm{softplus}(x))$ ~\citep{misra2020mish}, exhibits superior empirical results compared to \texttt{ReLU}, \texttt{Swish}, and \texttt{LReLU} in CIFAR-10 and ImageNet classification tasks. Fractional adaptive linear units \texttt{FALUs}~\citep{zamora2022fractional} incorporate fractional calculus principles into activation functions, thereby defining a diverse family of activation functions. It has demonstrated enhanced performance in image classification tasks, improving test accuracy. The \texttt{Seagull} activation function, introduced by ~\citep{gao2023data}, stands out as a customized activation function designed for applications in regression tasks featuring a partially exchangeable target function. 
It exhibits superiority in addressing the specific demands of regression scenarios. 

Overall, the above-mentioned activation functions are hard to be generalized across different domains, especially in industrial applications. Another problem is that the lack of theoretical analysis limits the acceptance of these activation functions in spite of their good performance. Therefore, it is necessary to verify an activation function with a good theoretical guarantee.

\subsection{Super-Expressive Activation Functions} \label{SuperExp AFs}

Numerous studies have explored new activation functions to make a fixed-size network achieve an arbitrary error, referred to as super-expressive activation functions. For example, \cite{MAIOROV199981} proposed an activation function to achieve this goal, but it lacks a closed form and is computationally impractical. Recently, \cite{yarotsky2021elementary} demonstrated that simple functions such as $(\texttt{sin}, \texttt{arcsin})$ can achieve super-expressiveness, although the relationship between the network size and the dimension was unclear. However, despite the above problems, \texttt{sin} has been proven to be effective in 3D neural network field, indicating the potential of super-expressiveness in neural networks \citep{ramirez2023deep}. \cite{2021Deep} proposed \texttt{EUAF}, showing that a network with \texttt{EUAF} requires only $\mathcal{O}(d^2)$ width and $\mathcal{O}(1)$ depth to achieve super-expressiveness. The potential of \texttt{EUAF} is demonstrated among simple experiments such as function approximation and Fashion-MNIST classification. They also explored the approximation of a neural network with three hidden layers which is named Floor-Exponential-Step (FLES) networks~\citep{Shen_2021}. The utilized floor function ($\left \lfloor x \right \rfloor $) can be recognized as an activation function with super-expressiveness~\citep{yarotsky2021elementary}. In a word, these super-expressive activation functions play a theoretically pivotal role in endowing models with the universal approximation property for all continuous functions. However, previous research either lacked experiments or only included simple ones, leaving it unknown whether these super-expressive functions are practically valuable. 

%%%%%%%%%%%%%%%%%%%%%%%%%%%%%%%%%%%%%%%%%%%%%%
%%%%%%%%%%%%%%%%%%%%%%%%%%%%%%%%%%%%%%%%%%%%%%
%%%%%%%%%% Section by Shijun

% \clearpage
%%%%%%%%%%%%%%%%%%%%%%%%%%%%%%%%%%%%%%%%%%%%%%
%%%%%%%%%% Section by Shijun
\section{Enriching the Family of Super-expressive Activation Functions}
% \subsection{Non-trivial Generalizations of EUAF}
\label{sec:generalization:EUAF}

In this section, we aim to significantly expand the scope of \texttt{EUAF} activation function by introducing a comprehensive collection of activation functions, each with approximation properties akin to those of \texttt{EUAF}. For simplicity, let $\nnOneD[]{\varrho}{N}{L}{\bbR^d}{\bbR^n}$ denote the set of neural networks $\phi:\bbR^d\to\bbR^n$ that can be represented by $\varrho$-activated networks, with a maximum width of $N$ and a maximum depth of $L$. Let $\scrA$ represent the set of all super-expressive activation functions $\varrho:\bbR\to\bbR$, which satisfy the following conditions:
\begin{itemize}
    \item There exists an interval $(\alpha,\beta)$ with $\alpha<\beta$ where $\varrho$ is real analytic and  non-polynomial on  $(\alpha,\beta)$.
    \item  There exists a fixed-size $\varrho$-activated network $\phi$ that can reproduce a triangle-wave function on $[0,\infty)$, i.e., 
    \[\tn{$\phi(x)=\big|x-2\big\lfloor \tfrac{x+1}{2}\big\rfloor\big|$ \quad  $ \forall  x\in [0,\infty)$.}\]
\end{itemize}
We denote $\barscrA$ as the ``closure" of $\scrA$. This means a function $\varrho$ is in $\barscrA$ if and only if, for any $A > 0$ and $\eps > 0$, there exists a $\varrho_\eps \in \scrA$ such that:
\begin{equation*}
    |\varrho_\eps(x)-\varrho(x)|<\eps\quad \forall  x\in [-A,A].
\end{equation*}

% \begin{tcboxThm}
\begin{theorem}
    \label{thm:main}
        Given any $\varrho\in \barscrA$, the hypothesis space
        $$\nnOneD[\big]{\varrho}{\calO(d^2)}{\calO(1)}{\bbR^d}{\bbR}$$
        is dense in $C([a,b]^d)$
        in terms of the supremum norm.
\end{theorem}
% \end{tcboxThm}

It is crucial to highlight that the constants in the $\calO(\cdot)$ notation in Theorem~\ref{thm:main} can be explicitly determined and depend only on the choice of $\varrho$.
The proof of Theorem~\ref{thm:main} will be provided later in this section. 

Before giving the proof, 
let us provide several examples in $\barscrA$. The first example, $\varrho_1\in \barscrA$, exhibits an S-shape and is defined as follows:
\begin{equation*}
    \varrho_1\coloneqq 
    \begin{cases}
        \frac{x}{1-x} &\tn{for}\ x\le  0,\\
        \tfrac{x}{1+x}+\frac{g(x)}{x^2+10} &\tn{for}\ x> 0, \\
    \end{cases}
\end{equation*}
where $g(x)=\big|x-2\big\lfloor \tfrac{x+1}{2}\big\rfloor\big|$ for any $x\in\bbR$.

The second example, $\varrho\in \barscrA$, resembles the \texttt{ReLU} activation function and is defined as follows:
\begin{equation*}
    \varrho_2\coloneqq 
    \begin{cases}
        0 &\tn{for}\  x\le 0, \\
        x+\frac{g(x)}{x+1} &\tn{for}\  x>0. \\
    \end{cases}
\end{equation*}

\begin{figure}[ht] 
\centering
\subfloat{
\includegraphics[width=0.31\linewidth]{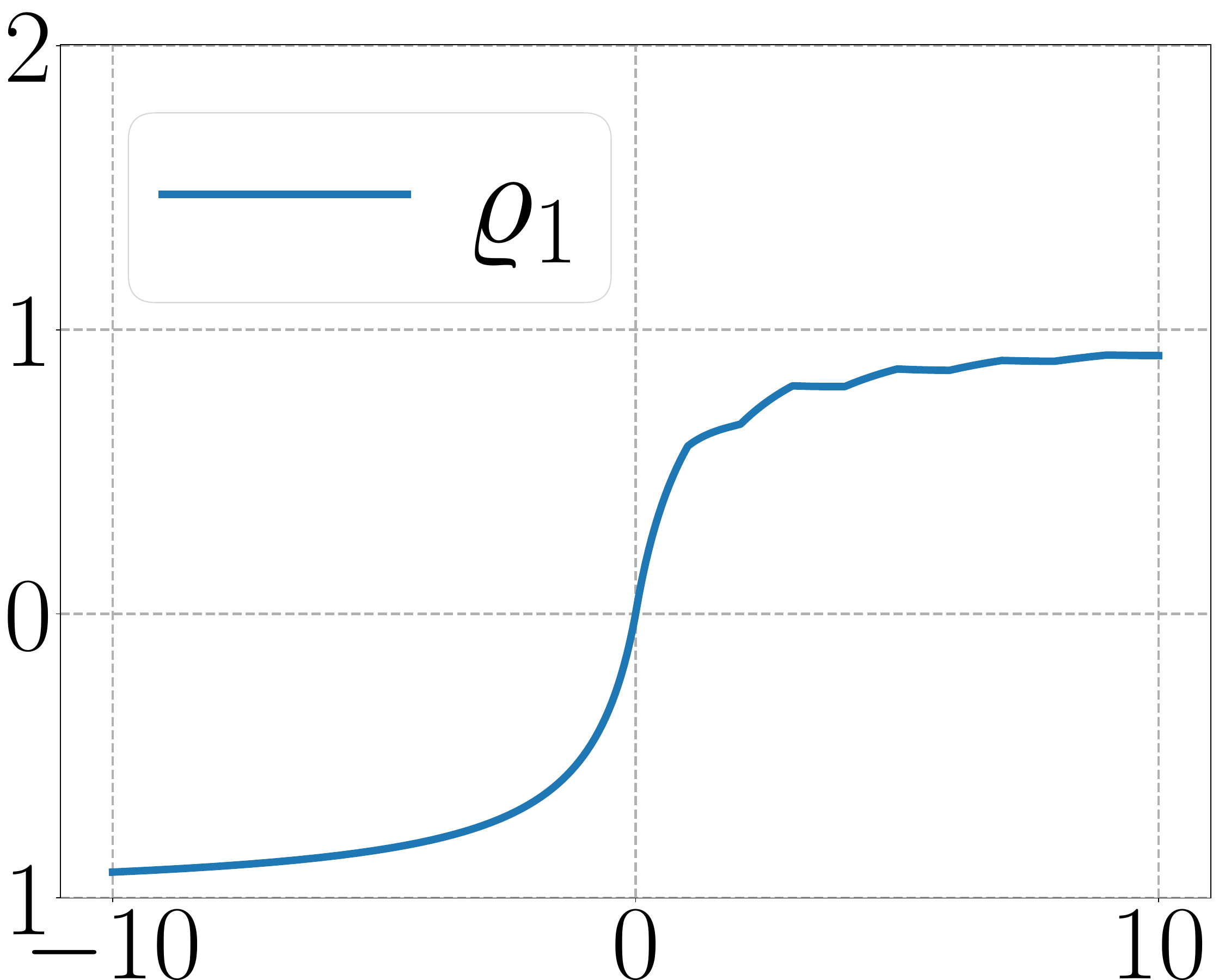}
% \label{fig:Loss for Resnet18 4Relu+1Mayactivation}
}
\subfloat{
\includegraphics[width=0.31\linewidth]{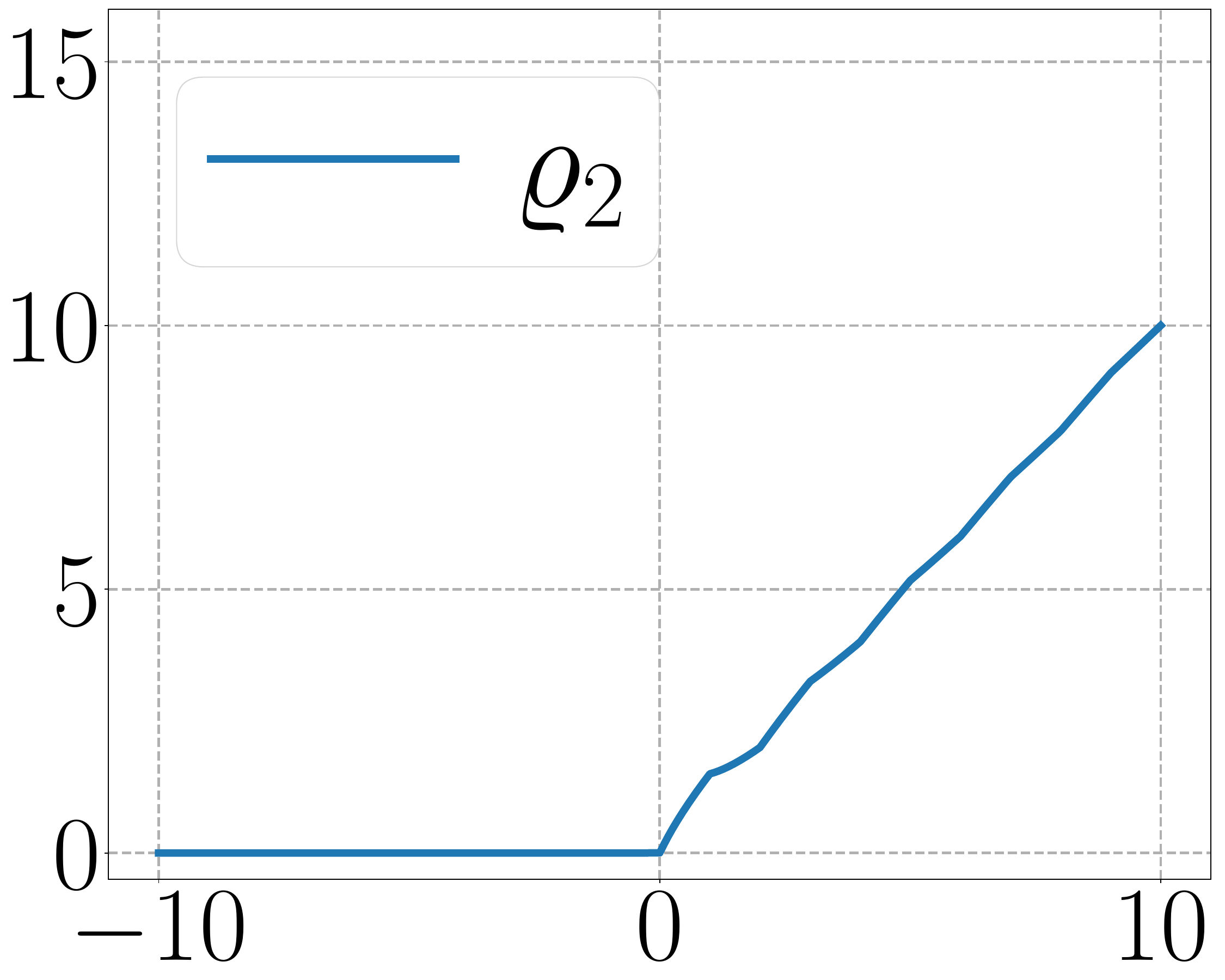}
% \label{fig:Loss for Resnet18 4Relu+1Mayactivation}
}
\subfloat{
\includegraphics[width=0.31\linewidth]{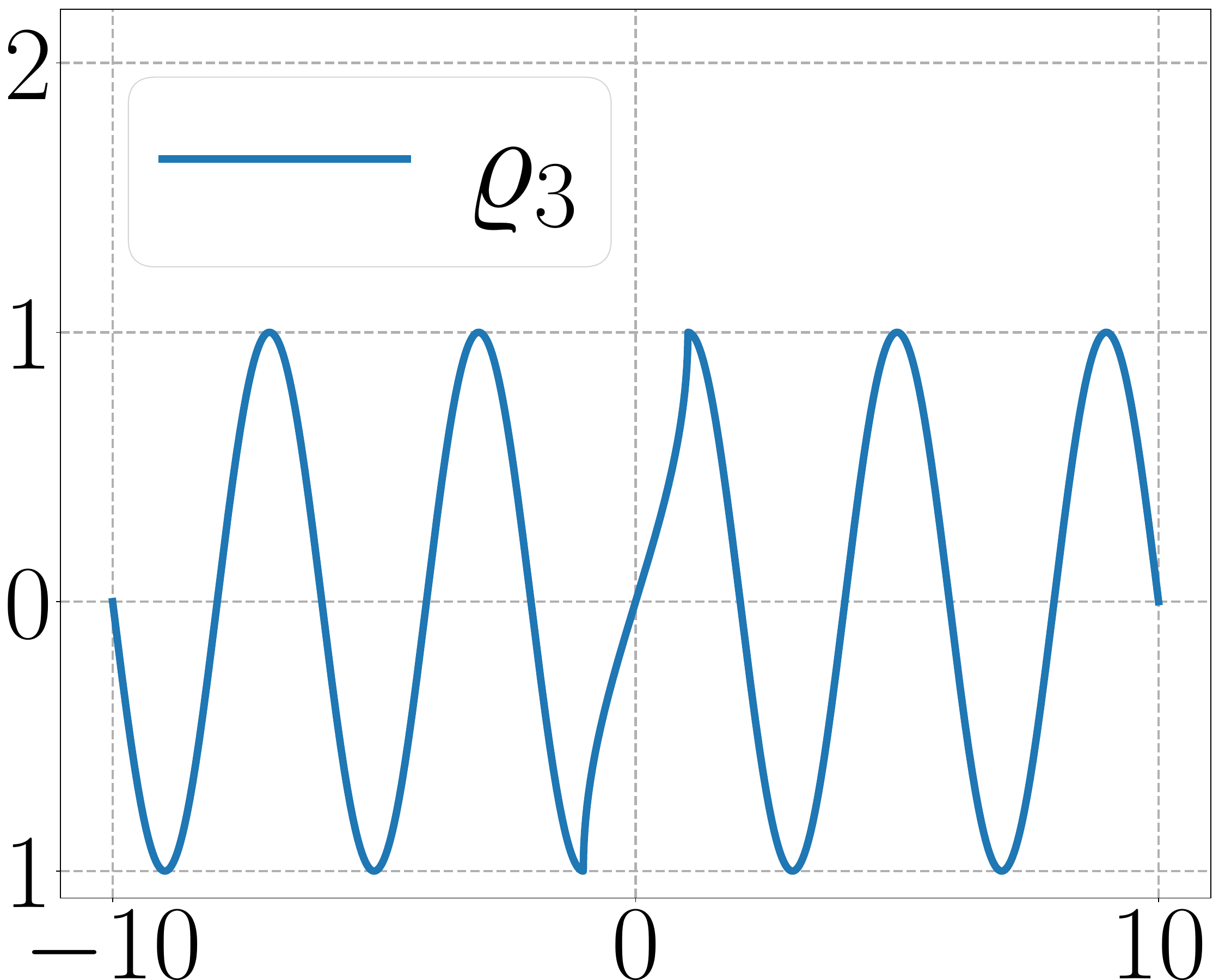}
% \label{fig:Loss for Resnet18 4Relu+1Mayactivation}
}
\caption{Illustrations of $\varrho_1$, $\varrho_2$, and $\varrho_3$. }
\label{fig:rho:123}
\vspace{-0.3cm}
\end{figure}

The third example, $\varrho_1\in\scrA\subseteq  \barscrA$, is defined as follows:
\begin{equation*}
    \varrho_3\coloneqq 
    \begin{cases}
        \tfrac{2}{\pi}\arcsin(x) & \tn{for}\ -1\le  x\le 1,\\
        \sin(\tfrac{\pi}{2}x) &\tn{for}\  |x|>1. \\
    \end{cases}
\end{equation*}
See  Figure~\ref{fig:rho:123} for visual representations of $\varrho_1$, $\varrho_2$, and $\varrho_3$.
% \begin{figure}[htbp!]        
% 	\centering          
% 		\includegraphics[width=0.95\linewidth]{rho1.pdf}
% 	\caption{An illustration of $\varrho_1$.}
% 	\label{fig:rho1}
% \end{figure}
% \begin{figure}[htbp!]        
% 	\centering          
% 		\includegraphics[width=0.95\linewidth]{rho2.pdf}
% 	\caption{An illustration of $\varrho_2$.}
% 	\label{fig:rho2}
% \end{figure}
% \begin{figure}[htbp!]        
% 	\centering          
% 		\includegraphics[width=0.95\linewidth]{rho3.pdf}
% 	\caption{An illustration of $\varrho_3$.}
% 	\label{fig:rho3}
% \end{figure}

Now, we will focus on proving the validity of Theorem~\ref{thm:main}.
% In order to establish Theorem~\ref{thm:main},
Given any any $f\in C([a,b]^d)$ and $\eps>0$, our goal is to construct $\phi\in \nnOneD[]{\varrho}{\calO(d^2)}{\calO(1)}{\bbR^d}{\bbR^n}$ such that
\begin{equation*}
    |\phi(\bmx)-f(\bmx)|<\eps\quad \forall  \bmx\in [a,b]^d.
\end{equation*}
Several concepts used to establish Theorem~\ref{thm:main} can be traced back to the research conducted by~\citep{2021Deep} and~\citep{yarotsky2021elementary}. The proof can be divided into three main steps as follows.
\begin{itemize}
    \item 
The primary objective of the first step is to create a neural network that effectively approximates the univariate function $f\in C([0,1])$ within a specific ``half" interval. 
% \begin{tcboxThm}
\begin{theorem}\label{thm:main:d=1:half}
    Given any $f\in C([0,1])$, $\varrho\in \barscrA$, $\varepsilon>0$, and $K\in\bbN$, suppose  for any $x_1,x_2\in[0,1]$, it holds that	
    \begin{equation}\label{eq:f:error:1/K}
		|f(x_1)-f(x_2)|<\varepsilon/2\  \tn{ if $|x_1-x_2|<1/K$.}
	\end{equation}
	Then there exists  $\phi\in \nnOneD[]{\varrho}{\calO(1)}{\calO(1)}{\bbR}{\bbR}$ such that 
	% $\|\phi\|_{L^\infty([0,1])}\le \|f\|_{L^\infty([0,1])}+1$ and
	\begin{equation*}
		|\phi(x)-f(x)|<\varepsilon\quad \tn{for any}\  x\in \bigcup_{k=0}^{K-1}\big[\tfrac{2k}{2K},\tfrac{2k+1}{2K}\big].
	\end{equation*}
\end{theorem}
% \end{tcboxThm}
\item
The second step's aim is to utilize the outcome of the first step,  Theorem~\ref{thm:main:d=1:half}, to build a network that effectively approximates the function $f\in C([a,b])$ within the entire interval $[a, b]$.
% \begin{tcboxThm}
\begin{theorem}\label{thm:main:d=1}
	Given any  $f\in C([a,b])$, $\varrho\in\barscrA$, and  $\varepsilon>0$,
	there exists $\phi\in \nnOneD[]{\varrho}{\calO(1)}{\calO(1)}{\bbR}{\bbR}$ such that
	\begin{equation*}
		|\phi(x)-f(x)|<\varepsilon\quad \tn{for any $x\in [a,b]$.}
	\end{equation*}
\end{theorem}
% \end{tcboxThm}
\item
The ultimate objective of the final step is to generalize the one-dimensional findings described in Theorem~\ref{thm:main:d=1} to the multi-dimensional scenario. To achieve this, we will utilize Kolmogorov's superposition theorem (KST) \citep{Kol1957}, summarized in Theorem~\ref{thm:kst}. It is important to note that the target function $f\in C([a, b]^d)$ can be appropriately rescaled to facilitate the application of KST.
% \begin{tcboxThm}
\begin{theorem}[KST]
	\label{thm:kst}
	There exist continuous functions $h_{i,j}\in C([0,1])$ for $i=0,1,\cdots,2d$ and $j=1,2,\cdots,d$ such that any continuous function $f\in C([0,1]^d)$ can be represented as 
	\begin{equation*}
		f(\bmx)=\sum_{i=0}^{2d}  g_i \Big(\sum_{j=1}^d h_{i,j}(x_j)\Big)
  % \quad \forall \tn{ $\bmx=(x_1,\cdots,x_d)\in [0,1]^d$,}
	\end{equation*}
 for any $\bmx=(x_1,\cdots,x_d)\in [0,1]^d$,
	where $g_i:\bbR\to\bbR$ is a continuous function for each $i\in \{0,1,\cdots,2d\}$.
\end{theorem}
% \end{tcboxThm}
\end{itemize}
We observe that it is sufficient to demonstrate the case where $\varrho\in \scrA$ rather than $\varrho\in \barscrA$, aided by the following lemma.
% \begin{tcboxLem}
\begin{lemma}[Proposition 10 of \citep{JMLR:v25:23-0912}]
%\label{prop:activation:replace}
	Given two functions $\varrho,\tildevarrho:\bbR\to\bbR$ with $\tildevarrho\in C(\bbR)$, suppose for any $M>0$, there exists
	$\tildevarrho_\eta\in \nnOneD[\big]{\varrho}{\tildeN}{\tildeL}{\bbR}{\bbR}$ for each $\eta\in (0,1)$
%	\begin{equation*}
%		\tildevarrho_\eta\in \nnOneD[\big]{\varrho}{\tildeN}{\tildeL}{\bbR}{\bbR}\quad \tn{for each $\eta\in (0,1)$}
%	\end{equation*}  
	such that
	\begin{equation*}
	\tildevarrho_\eta(x)\rightrightarrows \tildevarrho(x)\quad \tn{as}\  \eta\to 0^+\quad \tn{for any $x\in [-M,M]$.}
	\end{equation*}
	Assuming $\bmphi_{\tildevarrho}\in \nnOneD[\big]{\tildevarrho}{N}{L}{d}{n}$, for any $\eps>0$ and $A>0$, there exists
%	\begin{equation*}
%		\bmphi_{\varrho}\in \nn[\big]{\varrho}{\tildeN\cdot  N}{\  \tildeL\cdot  L}{d}{n}
%	\end{equation*}
	$\bmphi_{\varrho}\in \nnOneD[\big]{\varrho}{\tildeN\cdot  N}{\  \tildeL\cdot  L}{\bbR^d}{\bbR^n}$
	such that
	\begin{equation*}
		\big\|\bmphi_\varrho-\bmphi_\tildevarrho\big\|_{\sup([-A,A]^d)}<\eps.
	\end{equation*}
\end{lemma}
% \end{tcboxLem}
Now let's prove the utilized theorems.

\subsection{Proof of Theorem~\ref{thm:main:d=1:half}}
\begin{figure}[ht]%[htbp!]
    \centering
    \includegraphics[width=0.94775\linewidth]{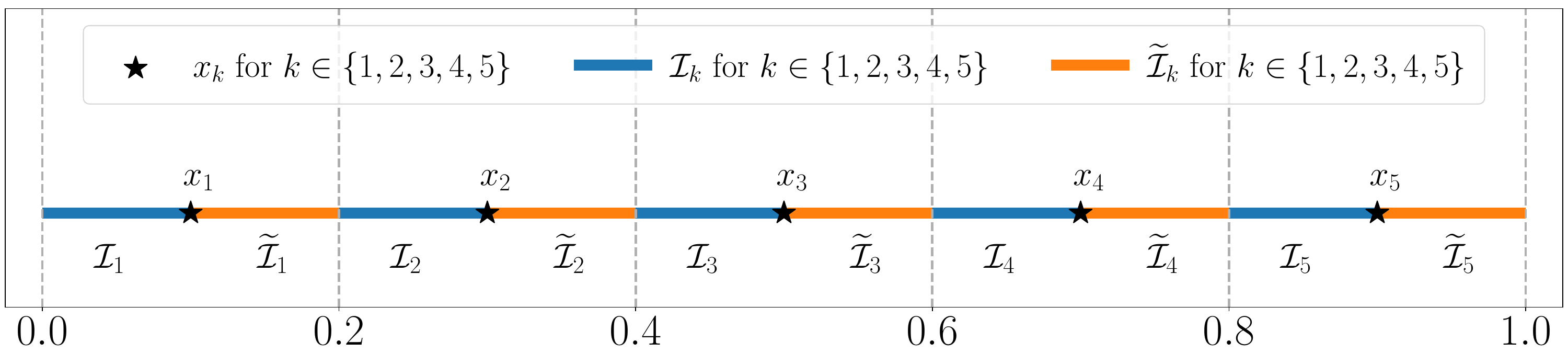}
    \caption{An illustration of $\calI_k$ and $\caltildeI_k$ for $k\in\{1,2,\cdots,K\}$ with $K=5$.}
    \label{fig:calI}
\end{figure}

% \mystep{1}{Approximation on a ``half'' interval.}
% For given $f\in C([0,1])$ and $\eps>0$, the goal of the first step is to achieve approximation on a ``half'' interval, i.e., constructing
%     $$\phi\in \nnOneD[]{\varrho}{\calO(1)}{\calO(1)}{\bbR}{\bbR}$$
%     such that
%         \begin{equation*}
%         |\phi(x)-f(x)|<\varepsilon\quad \tn{for any}\  x\in \bigcup_{k=0}^{K-1}\big[\tfrac{2k}{2K},\tfrac{2k+1}{2K}\big].
%     \end{equation*}
Partition $[0,1]$ into $2K$ small intervals $\calI_k$ and $\caltildeI_k$ for $k=1,2,\cdots, K$, i.e.,
\begin{equation*}
    \calI_k=\big[\tfrac{2k-2}{2K},\tfrac{2k-1}{2K}\big]\quad 
    \tn{and}\quad \caltildeI_k=\big[\tfrac{2k-1}{2K},\tfrac{2k}{2K}\big].
\end{equation*}
Clearly, $[0,1]=\bigcup_{k=1}^K(\calI_k\cup\caltildeI_k)$.
Let $x_k$ be the right endpoint of $\calI_k$, i.e., $x_k=\tfrac{2k-1}{2K}$ for $k=1,2,\cdots,K$.
 See an illustration of $\calI_k$, $\caltildeI_k$, and $x_k$ in Figure~\ref{fig:calI} for the case $K=5$.
Our objective is to construct $\phi\in \nnOneD[]{\varrho}{\calO(1)}{\calO(1)}{\bbR}{\bbR}$ to achieve accurate approximations of $f$ within $\calI_k$ for $k=1,2,\cdots,K$. It is not essential to consider the values of $\phi$ within $\caltildeI_k$ for all $k$. In other words, our focus is primarily on achieving accurate approximations within one ``half" of the interval $[0,1]$, which is the crucial element in our proof.

\begin{figure}[htbp!]
	\centering	\includegraphics[width=0.88\linewidth]{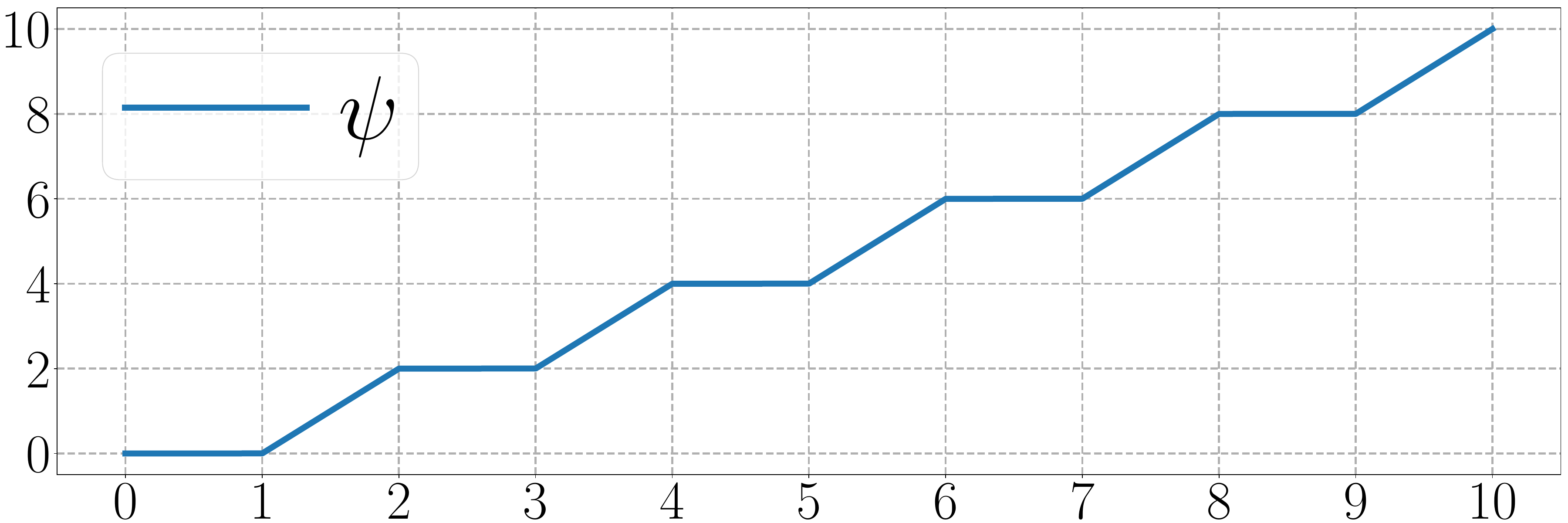}
	\caption{An illustration of $\psi$ on $[0,10]$.}
	\label{fig:psi}
\end{figure}

Define $\psi(x)\coloneqq  x-\sigma(x)$ for any $x\in \bbR$, where $\sigma\in \nnOneD[]{\varrho}{\calO(1)}{\calO(1)}{\bbR}{\bbR}$ with 
\begin{equation*}
    \sigma(x)=\big|x-2\big\lfloor \tfrac{x+1}{2}\big\rfloor\big|\quad \tn{ for $x\ge 0$.}
\end{equation*}
See Figure~\ref{fig:psi} for an illustration of $\psi$. 

It easy to verifty that
\begin{equation}\label{eq:psi:return:k}
	\psi(2Kx)/2+1=k\quad \tn{for any $x\in [\tfrac{2k-2}{2K},\tfrac{2k-1}{2K}]=\calI_k$.}
\end{equation}
We will make use of the two following lemmas to simplify our proof.
% \begin{tcboxLem}
\begin{lemma}[Lemma~23 of \cite{2021Deep}]
\label{lem:dense:stationary}
    Given any rationally independent numbers $a_1,a_2,\cdots,a_K$ for any $K\in\bbN^+$ and an arbitrary periodic function $g:\bbR\to\bbR$ with period $T$, i.e., $g(x+T)=g(x)$ for any $x\in\bbR$, assume there exist $x_1,x_2\in \bbR$ with $0<x_2-x_1<T$ such that $g$ is continuous on $[x_1,x_2]$. Then the following set 
    \begin{equation*}
        \Big\{\Big(u\cdot 
 g(wa_1)+v,\, \cdots,\,u\cdot g(wa_K)+v\Big): u,w,v\in\bbR
        \Big\}
    \end{equation*}
is dense in $\bbR^K$ provided that 
\begin{equation*}
    \tn{$\min\limits_{x\in [x_1,x_2]} g(x)<\max\limits_{x\in [x_1,x_2]} g(x)$.}
\end{equation*}
\end{lemma}
% \end{tcboxLem}

% \begin{tcboxLem}
\begin{lemma}
    \label{lem:rationally:independent}
	Given $K\in\bbN+$, suppose $\varrho$ is  real analytic and non-polynomial on an interval $(\alpha, \beta)$ with $\beta > \alpha$.
 Then there exists $w_0\in \big(-\tfrac{\beta-\alpha}{2K}, \tfrac{\beta-\alpha}{2K}\big)$ such that $\varrho\big(\frac{\alpha+\beta}{2} + k w_0 \big)$, for $\{k=1,2,\cdots,K\}$, are rationally independent.
\end{lemma}
% \end{tcboxLem}
\begin{proof}
We prove this lemma by contradiction. If it does not hold, then $\varrho\big(\frac{\alpha+\beta}{2} + k w \big)$, for $\{k=1,2,\cdots,K\}$, are rationally dependent for any $w\in \big(-\tfrac{\beta-\alpha}{2K}, \tfrac{\beta-\alpha}{2K}\big)=\calI$. That means, for any $w\in\calI$, there exists $\bmlambda=(\lambda_1,\cdots,\lambda_K)\in \bbQ^K\backslash\{\bmzero\}$ such that $\sum_{k=1}^K \lambda_k \varrho\big(\frac{\alpha+\beta}{2} + k w \big)=0$.
We observe that $\calI$ is uncountable and $\bbQ^K\backslash\{\bmzero\}$ is countable. It follows that there exists $\bmlambda=(\lambda_1,\cdots,\lambda_K)\in \bbQ^K\backslash\{\bmzero\}$ such that $\sum_{k=1}^K \lambda_k \varrho\big(\frac{\alpha+\beta}{2} + k w \big)=0$ for all $w$ in an uncountable subset of $\calI$. Then the real analyticity of $\varrho$ implies $\sum_{k=1}^K \lambda_k \varrho\big(\frac{\alpha+\beta}{2} + k w \big)=0$ for all $w\in \calI$.
By expanding $\sum_{k=1}^K \lambda_k \varrho\big(\frac{\alpha+\beta}{2} + k w \big)$ into the Taylor series at $w = 0$, we get the identity $\sum_{k=1}^K \lambda_k k^m = 0$ for each $m$ with $\frac{d^m\varrho}{dw^m}\left(\frac{\alpha+\beta}{2}\right) \neq 0$. Since $\varrho$ is non-polynomial on $(\alpha,\beta)\ni \tfrac{\alpha+\beta}{2}$, there are infinitely many $m$ with
$\frac{d^m\varrho}{dw^m}\left(\frac{\alpha+\beta}{2}\right) \neq 0$, implying $\sum_{k=1}^K \lambda_k k^m = 0$. This means $\bmlambda=(\lambda_1,\cdots,\lambda_K)=\bmzero$, a contradiction with $\bmlambda\in \bbQ^K\backslash\{\bmzero\}$. So we finish the proof of Lemma~\ref{lem:rationally:independent}.
\end{proof}

Now, let us return to the proof of Theorem~\ref{thm:main:d=1:half}.
We can employ Lemma~\ref{lem:rationally:independent} to produce a collection of rationally independent numbers. Specifically, there exists a value $w_0$ such that $a_1, a_2, \cdots, a_K$ are linearly independent, where each $a_k$ is defined as $a_k = \varrho\left(\tfrac{\alpha+\beta}{2} + kw_0\right)$.

Next,
define 
\begin{equation*}
    g(x)=\big|x-2\big\lfloor \tfrac{x+1}{2}\big\rfloor\big|\quad \tn{ for $x\in\bbR$.}
\end{equation*}
By Lemma~\ref{lem:dense:stationary}, there exists $u_1,w_1,v_1\in\bbR$ such that
\begin{equation*}
	\big|u_1\cdot  g(w_1 a_k)+v_1 -f(x_k)\big| <\varepsilon/2\quad \tn{for any $k$.}
\end{equation*}

Since $\sigma(x)=g(x)$ for any $x\ge 0$ and $g$ is periodic with period $2$, we can choose a sufficiently large $m_0\in\N$ such that
\begin{equation*}
\begin{split}
    	&\phantom{=\;\;}\big|u_1 \sigma(w_1 a_k+2m_0)+v_1-f(x_k)\big| \\
     &=\big|u_1 g(w_1 a_k+2m_0)+v_1-f(x_k)\big|
     \\
     &=\big|u_1 g(w_1 a_k)+v_1-f(x_k)\big|<\varepsilon/2,
\end{split}
\end{equation*}
for $k=1,2,\cdots,K$.
Define 
\begin{equation*}
    \phi(x)=u_1 \sigma\bigg(w_1 \varrho\Big(\tfrac{\alpha+\beta}{2} + (\tfrac{\psi(2kx)}{2}-1)w_0\Big)+2m_0\bigg)+v_1.
\end{equation*}
For any $x\in \calI_k$, we have
\begin{equation*}
    \begin{split}
        \phi(x)&=u_1 \sigma\bigg(w_1 \varrho\Big(\tfrac{\alpha+\beta}{2} + (\tfrac{\psi(2kx)}{2}-1)w_0\Big)+2m_0\bigg)+v_1
        \\
        &=u_1 \sigma\Big(w_1 \varrho\big(\tfrac{\alpha+\beta}{2} + kw_0\big)+2m_0\Big)+v_1
        \\
        &=u_1 \sigma\Big(w_1 a_k+2m_0\Big)+v_1,
    \end{split}
\end{equation*}
implying 
\begin{equation*}
|\phi(x)-f(x)|\le \underbrace{|\phi(x)-f(x_k)|}_{< \eps/2}
+
\underbrace{|f(x_k)-f(x)|}_{< \eps/2\tn{  by  } \eqref{eq:f:error:1/K}}
< \eps.
\end{equation*}
It follows that
\begin{equation*}
    |\phi(x)-f(x)|<\varepsilon\quad \tn{for any}\  x\in \bigcup_{k=0}^{K-1}\big[\tfrac{2k}{2K},\tfrac{2k+1}{2K}\big].
\end{equation*}
Moreover, we can easily verify
$\phi\in\nnOneD[]{\varrho}{\calO(1)}{\calO(1)}{\bbR}{\bbR}$.
So we finish the proof of Theorem~\ref{thm:main:d=1:half}.

% \begin{lemma}[Lemma~1 of \cite{yarotsky2021elementary}]    \label{lem:rationally:independent}
%     Let $\sigma$ be a real analytic function in an interval $(\alpha, \beta)$ with $\beta > \alpha$. Suppose that there is $N$ such that for all $w$ with sufficiently small absolute value, the values $\left(\sigma\left(\frac{\alpha+\beta}{2} + w_n\right)\right)^N_1$ are not rationally independent. Then $\sigma$ is a polynomial.
% \end{lemma}

\subsection{Proof of Theorem~\ref{thm:main:d=1} based on Theorem~\ref{thm:main:d=1:half}.}
% \mystep{2}{Approximation on a ``whole'' interval.}

We claim it suffices to prove the special case $[a,b]=[0,\tfrac{1}{2}]$ as this simplification readily extends to the broader scenario. To see this, we simply  introduce a linear function $\calL:[0,\tfrac{1}{2}]\to[a,b]$ by defining $\calL(x)=2(b-a)x+a$. 
The special case implies $f\circ \calL:[0,\tfrac{1}{2}]\to \bbR$ can be approximated by a network $\tildephi$ arbitrarily well. Then $\phi=\tildephi\circ \calL^{-1}$ can approximate $f:[a,b]\to\bbR$ well, as desired. 

We can continuously extend $f$ from $[0,\tfrac12]$ to $\bbR$ by setting $f(x)=f(0)$ if $x<0$ and $f(x)=f(\tfrac{1}{2})$ if $x>\frac{1}{2}$.
It follows from the uniform continuity of $f$ on $[-1,2]$ that there exists $K=K(f,\varepsilon)\in \bbN^+$ with $K\ge 2$ such that for any $x_1,x_2\in[-1,2]$,
\begin{equation*}
	|f(x_1)-f(x_2)|<\eps/10\quad \tn{if  $|x_1-x_2|<1/K$.}
\end{equation*}
For $i=1,2,3,4$, define
\begin{equation*}
	f_{i}(x)\coloneqq f\big(x-\tfrac{i}{4K}\big)\quad\tn{for any $x\in[0,1]$.}
\end{equation*}
Then, for $i=1,2,3,4$ and $x_1,x_2\in[0,1]$, we have
\begin{equation*}
	|f_i(x_1)-f_i(x_2)|<\eps/10=\tildeeps/2 \quad \tn{if $|x_1-x_2|<1/K$,}
\end{equation*}
where $\tildeeps=\eps/5$.
% which means we can apply Theorem~\ref{thm:main:d=1:half} to $f_i\in C([0,1])$.
For each $i\in \{1,2,3,4\}$, by Theorem~\ref{thm:main:d=1:half}, there exists  $\phi_i\in \nnOneD[]{\varrho}{\calO(1)}{\calO(1)}{\bbR}{\bbR}$ such that 
% \begin{equation*}
%     \|\phi_i\|_{L^\infty([0,1])}\le \|f_i\|_{L^\infty([0,1])}+1\le \|f\|_{L^\infty([-1,1])}+1
% \end{equation*}
% and
\begin{equation*}
	\big|\phi_i(x)-f_i(x)\big|<\widetilde{\varepsilon}=\varepsilon/5\quad \tn{for any}\  x\in \bigcup_{k=0}^{K-1}\big[\tfrac{2k}{2K},\tfrac{2k+1}{2K}\big].
\end{equation*}

\begin{figure}[htbp!]
	\centering	\includegraphics[width=0.82\linewidth]{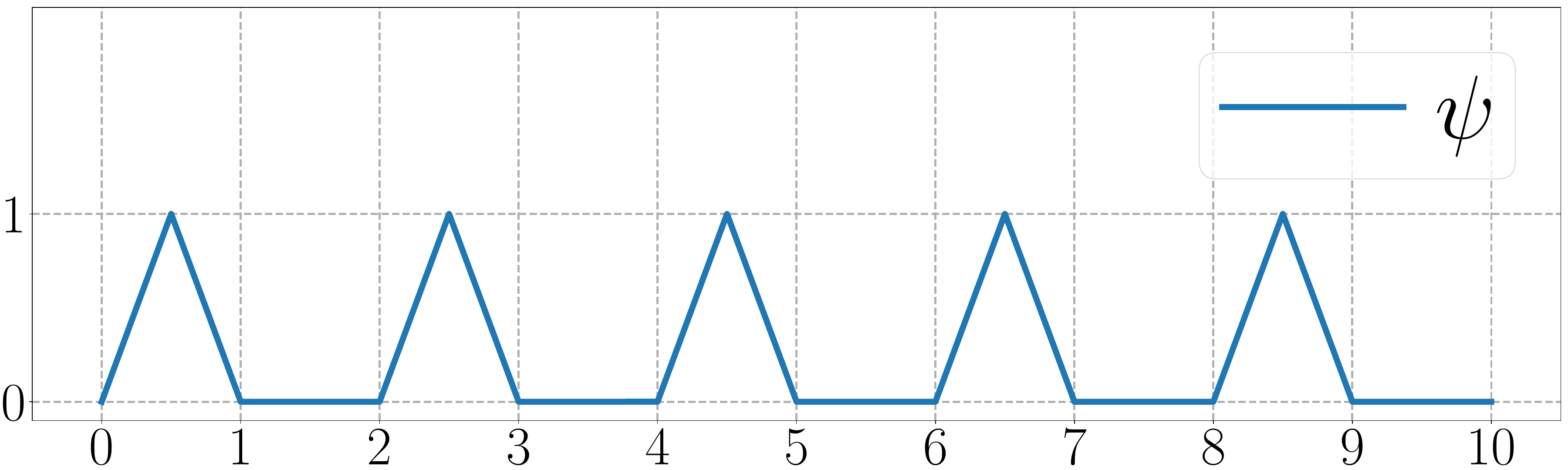}
	\caption{An illustration of $\psi$ on $[0,2K]$ for $K=5$.}
	\label{fig:psi:0:2K}
\end{figure}

Define 
\begin{equation*}
	\psi(x)=\sigma\big(x+1-\sigma(x+1)\big)\quad\tn{for any $x\in\bbR$,}
\end{equation*}
where $\sigma\in \nnOneD[]{\varrho}{\calO(1)}{\calO(1)}{\bbR}{\bbR}$ with 
\begin{equation*}
    \sigma(x)=\big|x-2\big\lfloor \tfrac{x+1}{2}\big\rfloor\big|\quad \tn{ for $x\ge 0$.}
\end{equation*}
See an illustration of $\psi$  on $[0,2K]$ for $K=5$ in Figure~\ref{fig:psi:0:2K}. 
%Note that $\sum_{i=1}^4 \psi(x+\tfrac{i}{4})=1$ for any $x\ge 0$.

Clearly, $0\le \psi(2Kx)\le 1$ for any $x\in [0,1]$, from which we deduce
\begin{equation*}
	\Big|\big(\phi_i(x)-f_i(x)\big)\psi(2Kx)\Big|
 % \le\big|\phi_i(x)-f_i(x)\big|
 <\varepsilon/5\quad \forall\  x\in \bigcup_{k=0}^{K-1}\big[\tfrac{2k}{2K},\tfrac{2k+1}{2K}\big].
\end{equation*}
Observe that $\psi(y)=0$ for $y\in \bigcup_{k=0}^{K-1} [{2k+1},{2k+2}]$, which implies 
\begin{equation*}
	\psi(2Kx)=0\quad \tn{for any $x\in \bigcup_{k=0}^{K-1} [\tfrac{2k+1}{2K},\tfrac{2k+2}{2K}]$}.
\end{equation*}
Subsequently, by the fact
\begin{equation*}
    [0,1]= \Bigg(\bigcup_{k=0}^{K-1}\big[\tfrac{2k}{2K},\tfrac{2k+1}{2K}\big]\Bigg)\bigcup \Bigg(\bigcup_{k=0}^{K-1}\big[\tfrac{2k}{2K},\tfrac{2k+1}{2K}\big]\Bigg),
\end{equation*}
we have
\begin{equation}\label{eq:phi:f:error:psi}
	\Big|\big(\phi_i(x)-f_i(x)\big)\psi(2Kx)\Big|<\varepsilon/5\quad \tn{for any $x\in[0,1]$}.
\end{equation}
%Therefore, for any $i=1,2,3,4$, we have
%\begin{equation*}
%	\Big|\phi_i(x)\psi(2Kx-\tfrac{i}{4K})-f_i(x)\psi(2Kx-\tfrac{i}{4K})\Big|\quad \tn{}.
%\end{equation*}

\begin{figure}[htbp!]        
	\centering
	\begin{minipage}{0.985\linewidth}
		\centering
		\subfloat{          
			\includegraphics[width=0.465\linewidth]{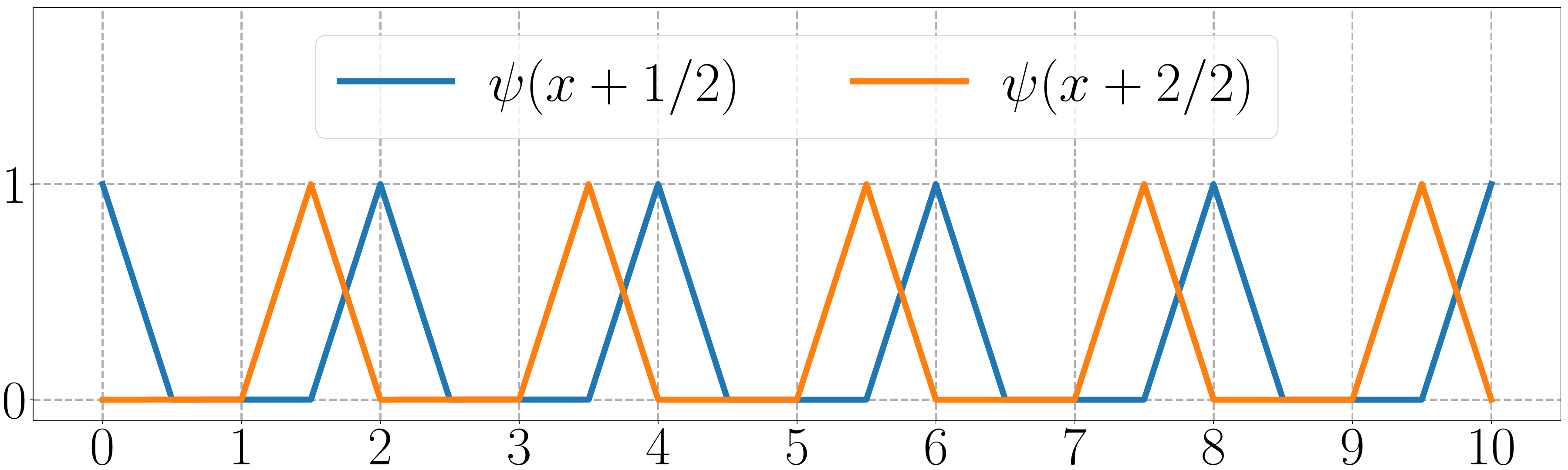}
}
		%		\begin{minipage}{0.07\textwidth}
		%			\,
		%		\end{minipage}
		\subfloat{         
			\includegraphics[width=0.465\linewidth]{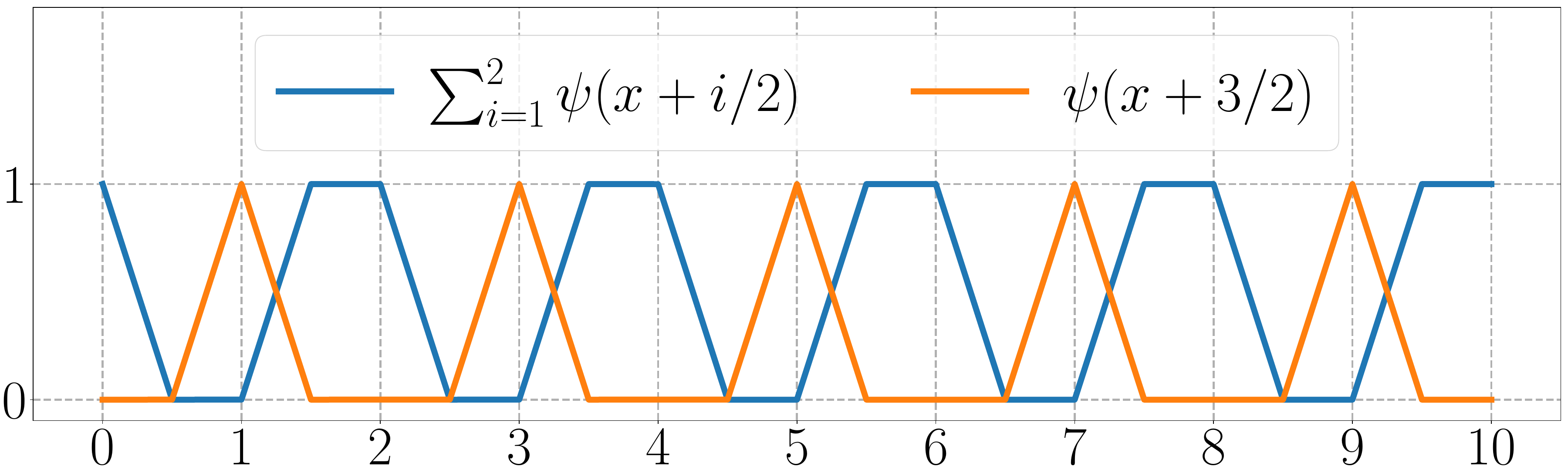}
%			\subcaption{}
}
		
		\vspace*{2pt}
\subfloat{           
		\includegraphics[width=0.465\linewidth]{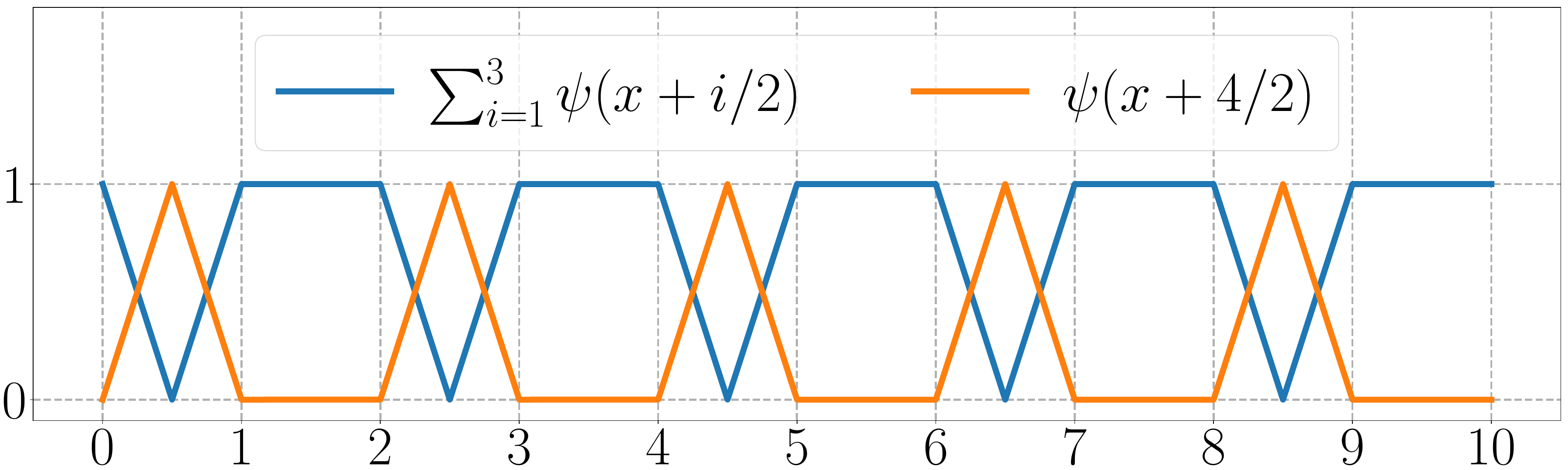}
%		\subcaption{}
}
\subfloat{         
	\includegraphics[width=0.465\linewidth]{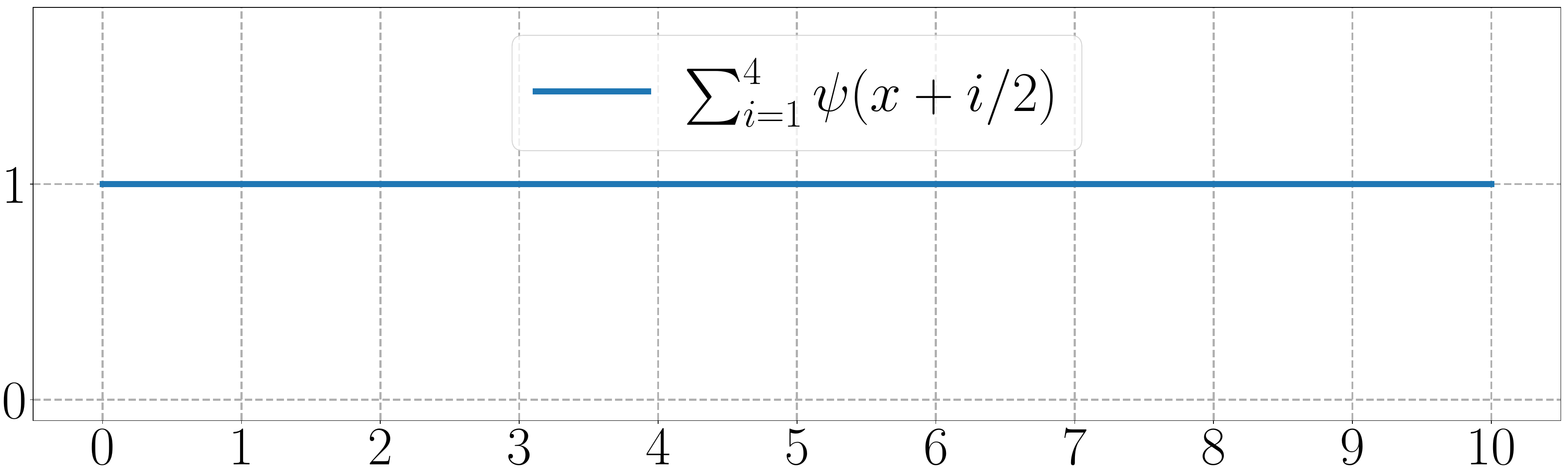}
	%		\subcaption{}
}
	\end{minipage}
	\caption{Illustrations of $\sum_{i=1}^{4}\psi(x+i/2)=1$ for any  $x\in [0,10]$.}
	\label{fig:psi1234}
\end{figure}

For each $i\in\{1,2,3,4\}$ and any $z\in[0,\tfrac12]
\subseteq [0,1-\tfrac{1}{K}]
\subseteq [0,1-\tfrac{i}{4K}]$, we have 
\begin{equation*}
    y_i=z+\tfrac{i}{4K}\in [\tfrac{i}{4K},1]\subseteq [0,1].
\end{equation*}
By bringing $x=y_i\in [0,1]$ into Equation~\eqref{eq:phi:f:error:psi}, we get
\begin{equation*}
% \label{eq:error:phi:component}
	\begin{split}
		&\phantom{=\;\;}\varepsilon/5
		> \Big|\big(\phi_i(y_i)-f_i(y_i)\big)\psi(2Ky_i)\Big|\\
  &=\Big|\phi_i(y_i)\psi(2Ky_i)-f_i(y_i)\psi(2Ky_i)\Big|\\
		&=\Big|\phi_i(z+\tfrac{i}{4K})\psi\big(2K(z+\tfrac{i}{4K})\big)-f_i(z+\tfrac{i}{4K})\psi\big(2K(z+\tfrac{i}{4K})\big)\Big|\\
		&=\Big|\phi_i(z+\tfrac{i}{4K})\psi\big(2Kz+\tfrac{i}{2}\big)-f(z)\psi\big(2Kz+\tfrac{i}{2}\big)\Big|
	\end{split}
\end{equation*}
for any $z\in [0,\tfrac12]$,
where the last equality comes from the fact that $f_i(x)=f(x-\tfrac{i}{4K})$ for any $x\in[0,1]\supseteq [\tfrac{i}{4K},1]$. Define 
\begin{equation*}
	\tildephi(x)\coloneqq\sum_{i=1}^4\phi_i(x+\tfrac{i}{4K})\psi\big(2Kx+\tfrac{i}{2}\big)\quad \tn{for any $x\in[0,\tfrac12]$.}
\end{equation*}
It is easy to verify that
$\sum_{i=1}^4 \psi\big(x+\tfrac{i}{2}\big)=1$ for any $x\ge 0$ based on the definition of $\psi$. See Figure~\ref{fig:psi1234} for illustrations.
It follows that $\sum_{i=1}^4 \psi\big(2Kz+\tfrac{i}{2}\big)=1$ for any $z\in [0,\tfrac12]$. 

Hence, for any $z\in [0,\tfrac12]$, 
% by Equation~\eqref{eq:error:phi:component}, 
we have
\begin{equation*}
	\begin{split}
		&\quad\;\big|\tildephi(z)-f(z)\big|
	\\	&=\Big|\sum_{i=1}^4\phi_i(z+\tfrac{i}{4K})\psi\big(2Kz+\tfrac{i}{2}\big)-f(z)\sum_{i=1}^4\psi\big(2Kz+\tfrac{i}{2}\big)\Big|\\
		&\le \sum_{i=1}^4\Big|\phi_i(z+\tfrac{i}{4K})\psi\big(2Kz+\tfrac{i}{2}\big)-f(z)\psi\big(2Kz+\tfrac{i}{2}\big)\Big|\\
  &< 4\cdot\frac{\varepsilon}{5}=\frac{4\eps}{5}.
	\end{split}
\end{equation*}
% That is, $|\phi(x)-f(x)|<\varepsilon$ for any $x\in[0,\tfrac12]$ as desired. 

To approximate $(x,y)\mapsto xy$ well, we define
\begin{equation*}
% \label{eq:psi:eps:z:def:new}
\Gamma_\delta(x,y)\coloneqq\tfrac{\varrho(x_0+\delta  x+\delta y)-\varrho(x_0+\delta y)-\varrho(x_0+\delta x)+\varrho(x_0)}{\delta^2\varrho^\dprime(x_0)}
\end{equation*}
for any $x,y\in \bbR$, where $\varrho^\dprime(x_0)\neq 0$. Clearly, $\Gamma_\delta(x,y)\to xy$ as $\delta\to 0$. Then we can define
\begin{equation*}	\phi_\delta(x)\coloneqq\sum_{i=1}^4\Gamma_\delta\Big(\phi_i(x+\tfrac{i}{4K}),\,\psi\big(2Kx+\tfrac{i}{2}\big)\Big)\quad \tn{ $\forall  x\in[0,\tfrac12]$.}
\end{equation*}
Clearly, $\phi_\delta\in \nnOneD[]{\varrho}{\calO(1)}{\calO(1)}{\bbR}{\bbR}$. Moreover, we can choose a sufficiently small $\delta_0>0$ such that
\begin{equation*}
    |\phi_{\delta_0}(x)-\tildephi(x)|<\eps/5\quad \tn{for any $x\in [0,\tfrac{1}{2}]$.}
\end{equation*}
By defining $\phi\coloneqq \phi_{\delta_0}\in \nnOneD[]{\varrho}{\calO(1)}{\calO(1)}{\bbR}{\bbR}$, we have
\begin{equation*}
    \begin{split}
        |\phi(x)-f(x)|
        % =|\phi_{\delta_0}(x)-f(x)|
        \le \underbrace{|\phi_{\delta_0}(x)-\tildephi(x)|}_{<\eps/5}
        +
        \underbrace{|\tildephi(x)-f(x)|}_{<4\eps/5}
        <\eps
    \end{split}
\end{equation*}
for any $x\in [0,\tfrac{1}{2}]$.
So we finish the proof of Theorem~\ref{thm:main:d=1}.

\begin{table*}[b]
\centering\footnotesize
\caption{A Brief Description of Three Image Datasets and Four Industrial Fault Diagnosis Datasets}
\label{Table: datasets}
\renewcommand{\arraystretch}{1.2}
\setlength{\tabcolsep}{5pt}
\scalebox{0.9}{\begin{tabular}{l  l }
\toprule
Dataset & Description  \\
\midrule
\href{https://www.cs.toronto.edu/~kriz/cifar.html}{CIFAR-10} & \makecell[l]{60,000 32×32 resolution RGB images in 10 categories (6,000 images per category) }\\
\midrule
\href{https://www.kaggle.com/c/tiny-imagenet}{Tiny ImageNet} & \makecell[l]{100,000  64×64 RGB images in 200 categories (500 for each category)}\\
\midrule
\href{https://www.image-net.org/}{ImageNet} & \makecell[l]{14,197,122 RGB images over 1,000 categories and 21,841 subcategories}\\
\midrule
Case Western Reserve University (\href{https://engineering.case.edu/bearingdatacenter/download-data-file}{CWRU} ) & \makecell[l]{2,400 vibration signals with 10 types of faults in drive end. Each signals has 1,024 samples} \\
\midrule
Power Quality Disturbance (\href{https://github.com/chachkes247/Power-Quality-Disturbances}{PQD}) & \makecell[l]{11200 voltage disturbance signals in 16 types, each disturbance signal at each fault has\\ additive white Gaussian noise} \\
\midrule
Motor Fault (\href{https://gitlab.com/power-systems-technion/motor-faults/-/tree/main}{MF}) & \makecell[l]{6 types of faults and each kind of fault has at least 290 samples ~\citep{sun2023public}} \\
\midrule
\makecell[l]{Electrical Fault Detection and Classification \\(\href{https://www.kaggle.com/datasets/esathyaprakash/electrical-fault-detection-and-classification}{EFDC})} & \makecell[l]{12,000 samples with 6 types of faults. Each sample has 6 features including the measured\\ line currents and voltages.} \\
\bottomrule
\end{tabular}}
\vspace{-0.3cm}
\end{table*}

\subsection{Proof of Theorem~\ref{thm:main} based on Theorem~\ref{thm:main:d=1} and KST.}
% \mystep{3}{Approximation on a hypercube.}

\begin{figure}[htbp!]
	\centering
	\includegraphics[width=0.98945\linewidth]{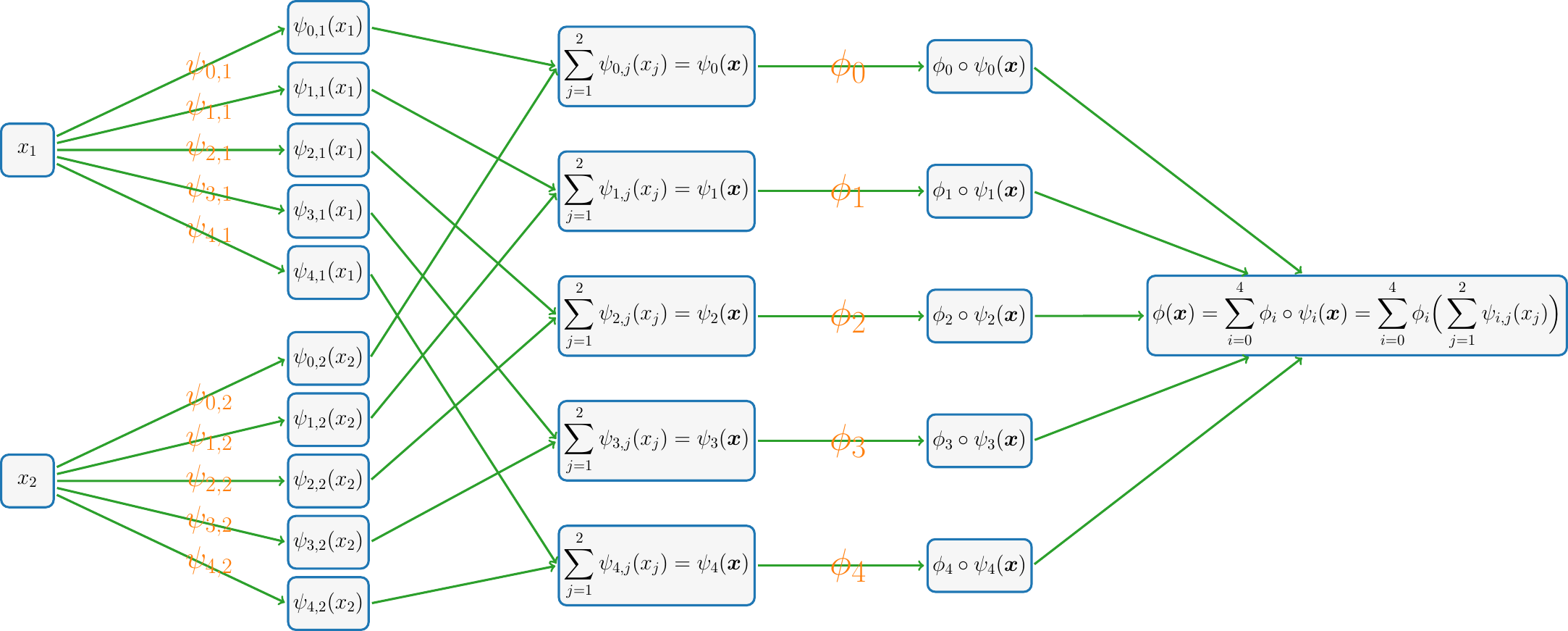}
	\caption{An illustration of the target network realizing  $\phi$ for any $\bmx\in [a,b]^d$ in the case of $d=2$. This network contains $(2d+1)d+(2d+1)=(d+1)(2d+1)$ sub-networks that realize $\psi_{i,j}$ and $\phi_i$ for $i=0,1,\cdots,2d$ and $j=1,2,\cdots,d$.}
	\label{fig:phi:decomposition}
\vspace{-0.2cm}
\end{figure}

We can safely assume that $[a, b] = [0, 1]$ since the general case can be readily extended by incorporating an affine map such as $\calL(a) = (b - a) x + a$.
Given any $f\in C([0,1]^d)$, by KST, 
there exist  $h_{i,j}\in C([0,1])$ and $g_i\in C(\bbR)$ for $i=0,1,\cdots,2d$ and $j=1,2,\cdots,d$ such that 
	\begin{equation*}
		f(\bmx)=\sum_{i=0}^{2d}  g_i \Big(\sum_{j=1}^d h_{i,j}(x_j)\Big)\quad \forall \tn{ $\bmx=(x_1,\cdots,x_d)\in [0,1]^d$.}
	\end{equation*}
 Choose a sufficiently large $A>0$, e.g.,  
 \begin{equation*}
     A=1+\sup\bigg\{\Big|\sum_{j=1}^d h_{i,j}(x_j)\Big|: i=0,1,\cdots,2d,\;\  \bmx\in [0,1]^d\bigg\}.
 \end{equation*}
 Then for any $\delta>0$, by  Theorem~\ref{thm:main:d=1}, there exist $\psi_{i,j},\phi_i\in \nnOneD[]{\varrho}{\calO(1)}{\calO(1)}{\bbR}{\bbR}$ such that
\begin{equation*}
	|g_{i}(t)-\phi_{i}(t)|<\delta\quad \tn{for any $t\in [-A,A]$}
\end{equation*}
and 
\begin{equation*}
	|h_{i,j}(t)-\psi_{i,j}(t)|<\delta\quad \tn{for any $t\in [0,1]$,}
\end{equation*}
 for $i=0,1,\cdots,2d$ and $j=1,2,\cdots,d$.
 By defining 
 \begin{equation*}
		\phi(\bmx)=\sum_{i=0}^{2d}  \phi_i \Big(\sum_{j=1}^d \psi_{i,j}(x_j)\Big)\quad \forall \tn{ $\bmx=(x_1,\cdots,x_d)\in \bbR^d$.}
	\end{equation*}
 we have $\phi\in \nnOneD[]{\varrho}{\calO(d^2)}{\calO(1)}{\bbR}{\bbR}$. See an illustration of the architecture of $\phi$ in Figure~\ref{fig:phi:decomposition}.
 Moreover, by choosing sufficiently small $\delta>0$, we can conclude that
 \begin{equation*}
     |\phi(\bmx)-f(\bmx)|<\eps\quad \forall  \bmx\in [0,1]^d,
 \end{equation*}
 which means we finish the proof of Theorem~\ref{thm:main}.

%%%%%%%%%%%%%%%%%%%%%%%%%%%%%%%%%%%%%%%%
%%%%%%%%%%%%%%%%%%%%%%%%%%%%%%%%%%%%%%%
\section{Experimental Results} \label{Experiments}

To further validate the efficacy of our activation functions, we evaluate \texttt{PEUAF} against a wide range of baseline activation functions, including \texttt{LReLU}~\citep{2015Empirical}, \texttt{PReLU}~\citep{he2015delving}, \texttt{Softplus}~\citep{7280459}, \texttt{ELU}~\citep{clevert2015fast}, \texttt{SELU} \citep{klambauer2017self}, \texttt{ReLU}~\citep{nair2010rectified} and \texttt{Swish}~\citep{ramachandran2017searching}. We conduct these comparisons across four industrial signal datasets and three image datasets (CIFAR-10 ~\citep{krizhevsky2009learning}, Fashion-MNIST~\citep{xiao2017fashionmnist} and ImageNet~\citep{deng2009imagenet}) using three distinct neural network architectures (LeNet-type~\citep{726791}, ResNet-18 ~\citep{2016Deep}, and VGG-16 ~\citep{simonyan2014very}).

As discussed in~\citep{2021Deep}, only a few neurons with super-expressive activation functions are required to approximate functions with arbitrary precision to avoid large generalization errors. However, implementing this in practical experiments is challenging. Therefore, our experiments primarily focus on exploring the feature patterns of \texttt{PEUAF} and determining how it contributes to improving test accuracy, instead of targeting 100\% test accuracy.

\subsection{Experimental Setups}
% The baselines are fixed to three blocks and each block has two 1D convolutional layers with 64 filters, $3 \times 1$ kernel size, and built-in activation function, one batch-normalization layer with suitable momentum, and one pooling layer with $3 \times 1$ pool size. Other advanced neural network typologies are fine-tuned according to datasets.
The datasets used in our experiments are briefly introduced in Table~\ref{Table: datasets}. For each experiment, we train the models with a batch size of 64 using the ``NAdam'' optimizer ~\citep{dozat2016incorporating}, with an initial learning rate of $0.01$. The learning rate decays with a factor of 0.2 if the accuracy change over 5 consecutive epochs is no more than $1\times 10^{-4}$. We set the number of epochs to 300 to ensure proper convergence. The baseline network structures employed in our experiments are introduced in Tables~\ref{tab:2} and~\ref{tab:Lenet}.

\begin{table}[htbp]
\centering\small
\caption{Baseline A for the CWRU, PQD, and MF datasets.}
\label{tab:2}
\renewcommand{\arraystretch}{1.2}
\resizebox{0.9\linewidth}{!}
{\begin{tabular}{ccc}
\toprule
Layer & Parameters & Activation \\ 
\midrule
1D-Convolution $(3 \times 1)$ & filter size=64, stride = 1 & \texttt{PEUAF} \\
1D-Convolution $(3 \times 1)$ & filter size=64, stride = 1 & \texttt{PEUAF} \\
\midrule
Batch-normalization (BN) & momentum=0.99, epsilon=0.001 & --\\
\midrule
Max-pooling  & pool size=$3 \times 1$,  stride = 1 & -- \\
1D-Convolution $(3 \times 1)$ & filter size=64, stride = 1 & \texttt{PEUAF} \\
1D-Convolution $(3 \times 1)$ & filter size=64, stride = 1 & \texttt{PEUAF} \\
\midrule
Batch-normalization (BN) & momentum=0.99, epsilon=0.001 & --\\
\midrule
Max-pooling  & pool size=$3 \times 1$,  stride = 1 & -- \\
1D-Convolution $(3 \times 1)$ & filter size=64, stride = 1 & \texttt{PEUAF} \\
1D-Convolution $(3 \times 1)$ & filter size=64, stride = 1 & \texttt{PEUAF} \\
\midrule
Batch-normalization (BN) & momentum=0.99, epsilon=0.001 & --\\
\midrule
% \multicolumn{3}{c}{Global-average-pooling}\\
Global-average-pooling &-- &-- \\
\midrule
Fully connected & size (chosen by tasks) & softmax \\
\bottomrule
\end{tabular}}
\vspace{-0.3cm}
\end{table}

\begin{table}[htbp]
\centering\small
\caption{Baseline B for the EFDC dataset.}
\label{tab:Lenet}
\renewcommand{\arraystretch}{1.2}
\resizebox{0.9\linewidth}{!}
{\begin{tabular}{ccc}
\toprule
Layer & Parameters & Activation \\ 
\midrule
1D-Convolution $(2 \times 1)$ & filter size=16, stride = 1 & \texttt{PEUAF} \\
\midrule
Batch-normalization (BN) & momentum=0.99, epsilon=0.001 & --\\
\midrule
Max-pooling  & pool size=$2 \times 1$,  stride = 1 & -- \\
1D-Convolution $(2 \times 1)$ & filter size=16, stride = 1 & \texttt{PEUAF} \\
\midrule
Batch-normalization (BN) & momentum=0.99, epsilon=0.001 & --\\
\midrule
Max-pooling  & pool size=$2 \times 1$,  stride = 1 & -- \\
\midrule
% \multicolumn{3}{c}{Flatten}\\
Flatten &-- &-- \\
\midrule
Fully connected &   size (chosen by tasks) & softmax \\
\bottomrule
\end{tabular}}
\vspace{-0.3cm}
\end{table}

\begin{figure}[ht] 
\centering
\includegraphics[width=0.95\linewidth]{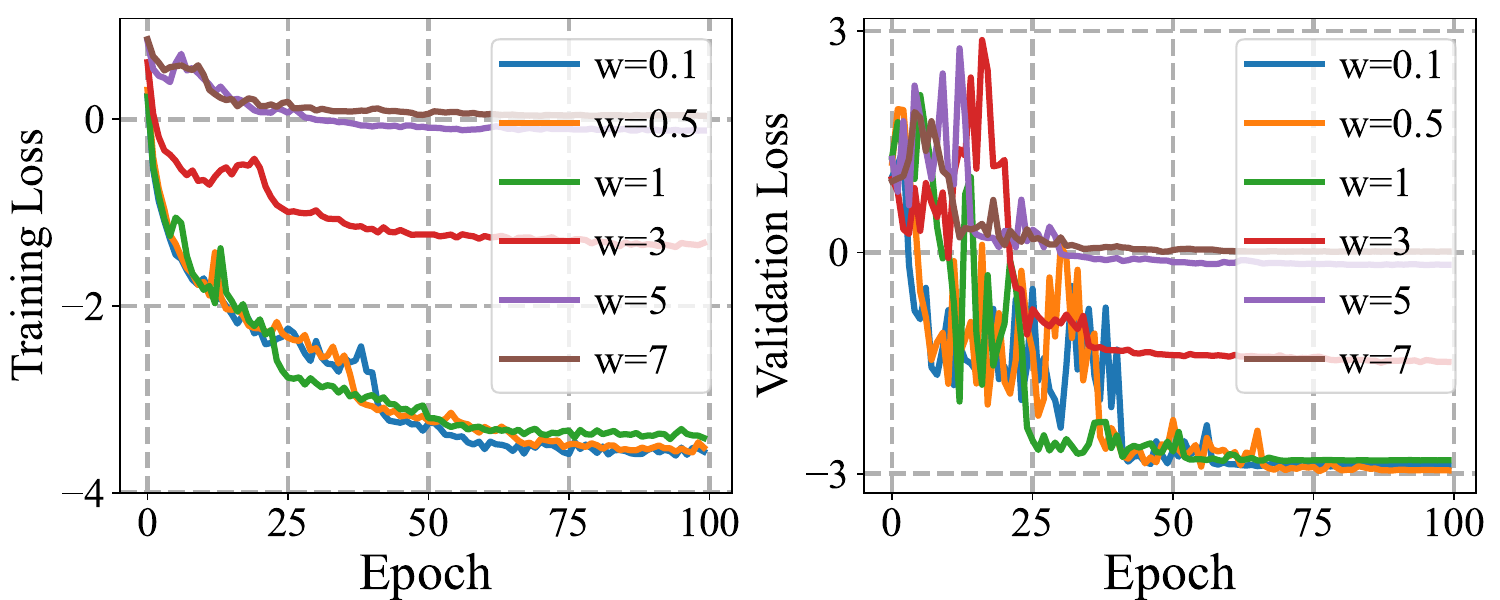}\\
% \label{fig:/PQD all Loss with different w}
% \subfloat[]{
% \includegraphics[width=0.85\linewidth]{figure/PQD Validation Loss with different alpha.pdf}
% \label{fig:PQD Validation Loss with different w}
% }\vspace{-0.3cm}
\caption{The training and validation loss with different $w$. (a) the training loss; (b) the validation loss}
\label{fig:Loss with different w}
\vspace{-0.3cm}
\end{figure}

The most critical hyperparameter is the range of the adaptive frequency $w$. To determine this, we conducted a classification experiment with different $w$ values on the PQD dataset ~\citep{0Open}, as illustrated in Figure~\ref{fig:Loss with different w}. The network structure used is Baseline A, a 1D convolutional neural network. To emphasize the discrepancies in outcomes, we employ a logarithmic transformation ($\log$) during the visualization of the loss function. Figure~\ref{fig:Loss with different w} shows the training and validation curves, while Table~\ref{Table:The accuracy of PQD classification with different w} provides the corresponding test accuracy. The table reveals two key points: First, when $w$ exceeds 1, the test accuracy drops significantly, indicating that higher frequencies pose challenges to the \texttt{PEUAF}'s ability to effectively extract PQD features. Second, when $w$ lies in the range of $[0, 1]$, the test accuracy consistently remains above 98\%. Therefore, we reasonably conclude that the frequency $w$ should be constrained within the range of $[0, 1]$.

\begin{table}[htbp!]
\centering\footnotesize
\caption{The accuracy of PQD classification with different $w$.}
\label{Table:The accuracy of PQD classification with different w}
\renewcommand{\arraystretch}{1.2}
\setlength{\tabcolsep}{5pt}
\scalebox{0.9}{\begin{tabular}{lccccccc }
 \toprule
$w$ & 0.1 & 0.5 & 1 & 3 & 5 & 7  \\
 \midrule
Accuracy & 98.25$\%$ & 98.30$\%$ & 98.04$\%$ & 93.66$\%$ & 75.26$\%$ & 53.39$\%$  \\
\bottomrule
\end{tabular}}
\end{table}

%二维卷积 ReLU和FEUAF混用

\subsection{Analysis Experiments} \label{1D}

In this section, we conduct experiments to show the characteristics of \texttt{PEUAF}. For the larger datasets (CWRU, PQD, and MF), we utilize the Baseline A in Table~\ref{tab:2}, while for the EFDC dataset, we use the Baseline B in Table~\ref{tab:Lenet}. Baseline B is smaller than Baseline A due to the smaller size of the EFDC dataset compared to CWRU, PQD, and MF. Our comparison focuses not only on the overall performance but also on the convergence behavior during the training process, fluctuations in the validation process, and a detailed mechanism analysis. 

\begin{table}[htbp]
\centering\footnotesize
\caption{Test accuracy in industrial fault diagnosis datasets.}
\label{Table:Test accuracy in industrial datasets}
\renewcommand{\arraystretch}{1.2}
\setlength{\tabcolsep}{5pt}
\scalebox{0.9}{\begin{tabular}{l l l l l c}
\toprule
Model & CWRU & PQD & MF & EFDC & Avg Rank  \\
\midrule 
\texttt{LReLU} & 99.58$\%$ & 98.12$\%$  & 98.82$\%$ & 84.24$\%$ & 4 \\
\texttt{PReLU} & 97.08$\%$ & 97.85$\%$  & 99.17$\%$ & 84.50$\%$& 7\\
\texttt{Softplus} & 95.00$\%$ & 97.95$\%$  & 98.06$\%$ &84.37$\%$& 8 \\
\texttt{ELU} & \textbf{100.00$\%$} & 98.15$\%$  &99.70$\%$  &83.10$\%$ & 2\\
\texttt{SELU} &99.17$\%$ & 98.12$\%$  &99.85$\%$  & 83.74$\%$ & 3\\
\texttt{ReLU} &99.58$\%$ & 97.89$\%$  &99.70$\%$  & 83.35$\%$& 5\\
Swish &99.16$\%$ &98.15$\%$   & 97.35$\%$ &84.63$\%$ & 6\\
\texttt{PEUAF} &\textbf{100.00$\%$} &\textbf{98.17$\%$}   &\textbf{100.00$\%$}  &\textbf{85.64$\%$} & \textbf{1}\\
\bottomrule
\end{tabular}}
% \vspace{-0.3cm}
\end{table}

\textbf{Performance}. Table~\ref{Table:Test accuracy in industrial datasets} summarizes the performance of several activation functions. All results are the average over three runs. On the EFDC dataset, \texttt{PEUAF} takes the lead by the largest margin, \textit{i.e.}, surpassing the second place \texttt{Swish} by over $1\%$. On the CWRU, dataset, \texttt{PEUAF} exhibits competitive performance compared to \texttt{Swish}, \texttt{ReLU}, \texttt{SELU}, \texttt{ELU}, and \texttt{LReLU}, while \texttt{PEUAF} outperforms \texttt{Softplus} and \texttt{PReLU} by 2\% and 5\%, respectively. On the PQD dataset, all activation functions achieve similar test accuracy. Lastly, on the MF dataset, \texttt{PEUAF} shows similar performance with \texttt{ReLU}, \texttt{SELU}, \texttt{ELU}, and \texttt{PReLU} but surpasses \texttt{Swish} and \texttt{Softplus}. Overall, \texttt{PEUAF} proves to be a competent activation function on four industrial fault diagnosis datasets. 

\begin{figure}[!t] 
% \vspace{-0.4cm}
\centering
\includegraphics[width=0.95\linewidth]{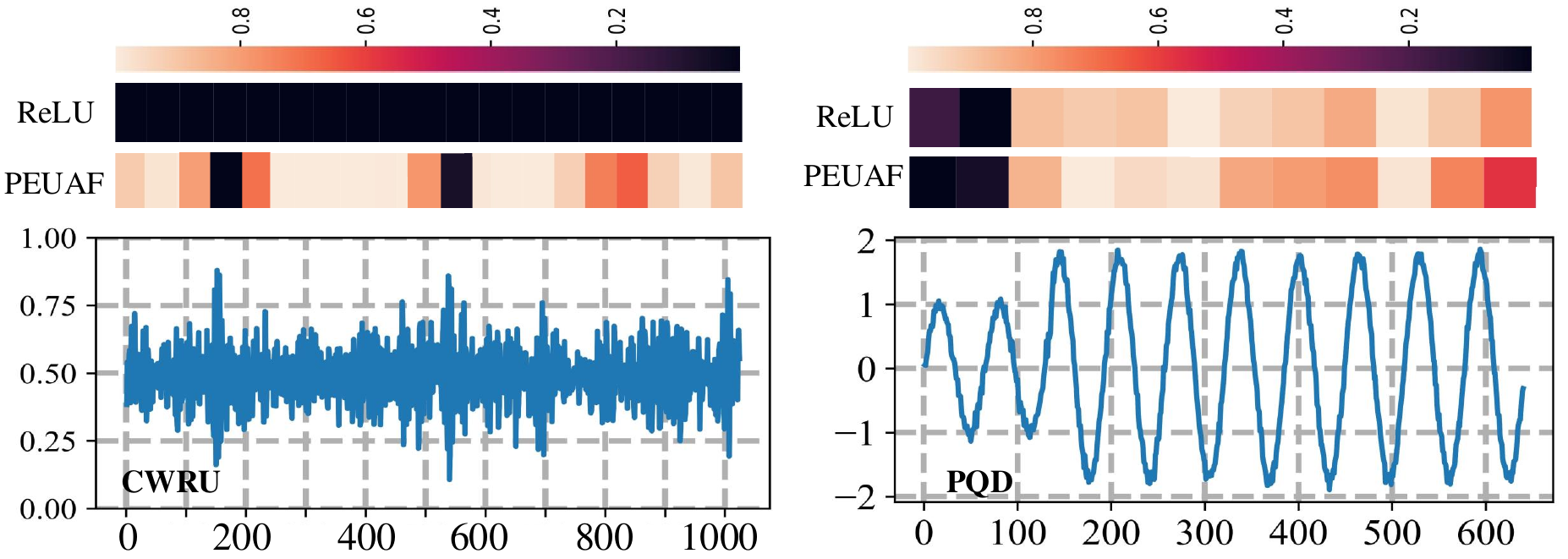}
% \subfloat[]{
% \includegraphics[width=0.95\linewidth]{figure/CWRU_ReLU_EUAF_Occlusion.pdf}
% \label{fig:CWRU_ReLU_EUAF_Occlusion}
% }~\vspace{-0.1cm}\\
% \subfloat[]{
% \includegraphics[width=0.95\linewidth]{ figure/PQD_ReLU_EUAF_Occlusion.pdf }
% \label{fig:PQD_ReLU_EUAF_Occlusion}
% }\vspace{-0.1cm} \\
% \subfloat[]{
% \includegraphics[width=0.95\linewidth]{
% figure/MF_ReLU_EUAF_Occlusion.pdf}
% \label{fig:MF_ReLU_EUAF_Occlusion}
% }~ \vspace{-0.1cm} \\
\caption{Occlusion experiments of Baselines using \texttt{PEUAF} and \texttt{ReLU} in CWRU and PQD datasets. It is seen that the baseline with \texttt{PEUAF} can better localize the fault when the original signal is noisy. }
\label{fig:Occlusion experimenst of Baselines in three datasets}
\vspace{-0.2cm}
\end{figure}

To further evaluate the effectiveness of \texttt{PEUAF}, we conducted occlusion experiments in two classic datasets: CWRU and PQD. For each dataset, the occluding sizes and strides were set to 100 and 50, respectively. The occluded pixels were all replaced by zeros. Based on the results in Figure~\ref{fig:Occlusion experimenst of Baselines in three datasets}, we observe that  \texttt{PEUAF} outperforms \texttt{ReLU} in locating faults. The experiments reveal two distinct levels of performances: (1) In the PQD dataset, \texttt{PEUAF} and \texttt{ReLU} show similar performance in accurately detecting and localizing faults, as illustrated in Figure~\ref{fig:Occlusion experimenst of Baselines in three datasets}. This can be attributed to the favorable condition within the PQD dataset, characterized by its low signal-to-noise ratio, contributing to the successful faults localization. However, such ideal conditions are rare in real-world scenarios. (2) In contrast, in the CWRU dataset, \texttt{PEUAF} significantly outperforms \texttt{ReLU} as shown in Figure~\ref{fig:Occlusion experimenst of Baselines in three datasets}. Despite that both \texttt{PEUAF} and \texttt{ReLU} achieve commendable test accuracy, \texttt{ReLU} tends to capture more holistic features instead of locating the real fault, which makes the outputs less reliable. Conversely, \texttt{PEUAF} effectively locates faults even in the presence of noise interference, offering valuable insights into the timing and severity of fault occurrences, as indicated by the occlusion experiments.

% appendix \ref{sectionA} and \ref{sectionB}, the \texttt{PEUAF} outperforms \texttt{ReLU}. The results reveal two distinct performance. Taking the baselines as examples, the first level is that \texttt{PEUAF} and \texttt{ReLU} have analogous performance, as illustrated in Figure~\ref{fig:Occlusion_exp for PQD Baseline FEUAF}, ~\ref{fig:Occlusion_exp for PQD Baseline ReLU}, ~\ref{fig:Occlusion_exp for MF Baseline FEUAF}, and ~\ref{fig:Occlusion_exp for MF Baseline ReLU}. In this level, both activation functions facilitate the network to accurately detect and localize faults. This achievement can be attributed to the favorable conditions within the PQD dataset, characterized by its low signal-to-noise ratio, and the absence of noise within the MF dataset, which contribute to the successful localization of faults. However, such situation is rare in real-world field data. The second level of performance  is \texttt{PEUAF} outperforms  \texttt{ReLU} as exemplified in Figure~\ref{fig:Occlusion_exp for CWRU Baseline FEUAF} and \ref{fig:Occlusion_exp for CWRU Baseline ReLU}. Despite both \texttt{PEUAF} and \texttt{ReLU} achieving commendable test accuracy, it is apparent that \texttt{ReLU} prefers to capture a more holistic features instead of locating the real fault which makes the outputs untrustworthy. Conversely, \texttt{PEUAF} effectively locates faults even in the presence of noise interference, offering valuable insights into the timing and severity of fault occurrences, as indicated by the occlusion experiments.

% The training and validation loss, and the corresponding occlusion experiments are separately shown in Figure~\ref{fig:Loss of Baselines in three datasets}, Figure~\ref{fig:Occlusion experimenst of Baselines in three datasets}, Appendix~\ref{sectionA}, and Appendix~\ref{sectionB}. Table~\ref{Table:Test accuracy in three datasets} presents all test accuracy of the four typologies in the three datasets.

% \begin{table}[htbp]
% \centering\footnotesize
% \caption{Test accuracy in industrial datasets.}
% \label{Table:Test accuracy in industrial datasets}
% \renewcommand{\arraystretch}{1.2}
% \setlength{\tabcolsep}{5pt}
% \begin{tabular}{l l l l l l l}
% \toprule
% Model & \multicolumn{2}{c}{CWRU} & \multicolumn{2}{c}{PQD} & \multicolumn{2}{c}{MF}  \\
% \midrule
% -- & \texttt{PEUAF} & ReLU & \texttt{PEUAF} & ReLU & \texttt{PEUAF} & ReLU  \\

% \midrule
% Baseline & 100$\%$ & 99.37$\%$  & 98.17$\%$  & 97.89$\%$  &100$\%$ & 99.59$\%$ \\
% ResNet & 100$\%$& 100$\%$  & 98.21$\%$  & 98.08$\%$  & 100$\%$ & 99.85$\%$ \\
% InceptionNet & 100$\%$ & 100$\%$  & 98.15$\%$  & 98.13$\%$  & 100$\%$ & 100$\%$\\
% VGG & 100$\%$ & 100$\%$  &  97.72$\%$  & 97.72$\%$  & 100$\%$ & 100$\%$\\
% \bottomrule
% \end{tabular}
% \end{table}

\begin{figure}[ht] 
\centering
\includegraphics[width=1\linewidth]{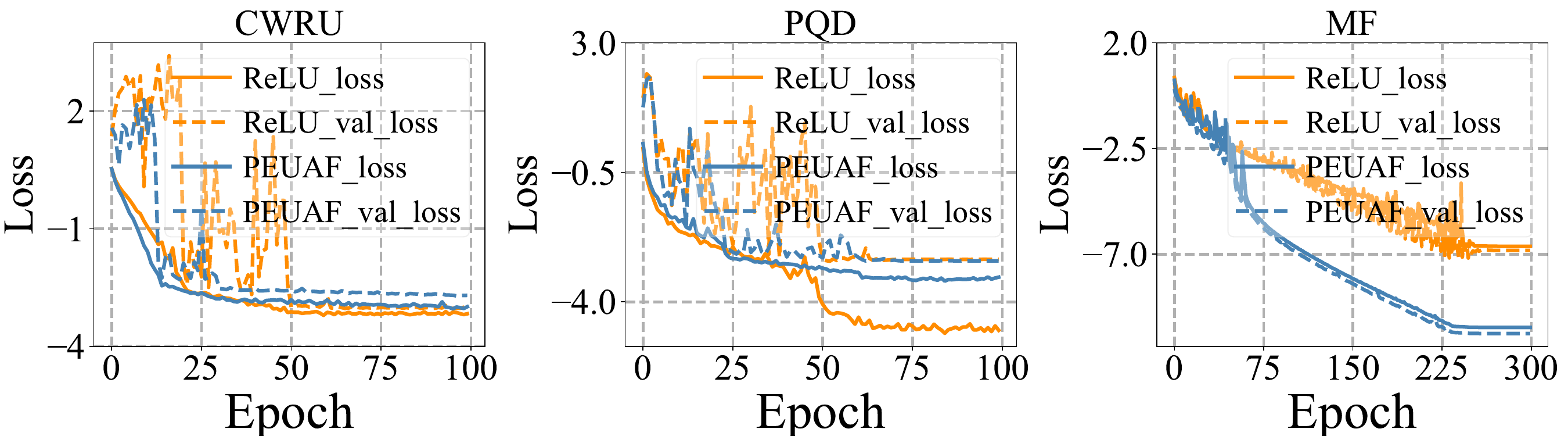}
% \subfloat[]{
% \includegraphics[width=0.43\linewidth]{figure/Loss for CWRU Baseline ReLU.pdf}
% \label{fig:Loss for CWRU Baseline ReLU}
% }\vspace{-0.1cm}\\
% \subfloat[]{
% \includegraphics[width=0.45\linewidth]{figure/Loss for PQD Baseline FEUAF.pdf}
% \label{fig:Loss for PQD Baseline FEUAF}
% }~\vspace{-0.1cm}
% \subfloat[]{
% \includegraphics[width=0.45\linewidth]{figure/Loss for PQD Baseline ReLU.pdf}
% \label{fig:Loss for PQD Baseline ReLU}
% }\vspace{-0.1cm}\\
% \subfloat[]{
% \includegraphics[width=0.45\linewidth]{figure/Loss for MF Baseline FEUAF.pdf}
% \label{fig:Loss for MF Baseline FEUAF}
% }~\vspace{-0.05cm}
% \subfloat[]{
% \includegraphics[width=0.45\linewidth]{figure/Loss for MF Baseline ReLU.pdf}
% \label{fig:Loss for MF Baseline ReLU}
% }\vspace{-0.05cm}\\
% \subfloat[]{
% \includegraphics[width=0.3\linewidth]{figure/Loss for CWRU Baseline ReLU.pdf}
% \label{fig:Loss for CWRU Baseline ReLU}
% }~
% \subfloat[]{
% \includegraphics[width=0.3\linewidth]{figure/Loss for PQD Baseline ReLU.pdf}
% \label{fig:Loss for PQD Baseline ReLU}
% }~
% \subfloat[]{
% \includegraphics[width=0.3\linewidth]{figure/Loss for MF Baseline ReLU.pdf}
% \label{fig:Loss for MF Baseline ReLU}
% }\\
\caption{Training dynamics of Baselines using \texttt{PEUAF} and \texttt{ReLU} in three large datasets. }
\label{fig:Loss of Baselines in three datasets}
\vspace{-0.3cm}
\end{figure}

\textbf{Convergence}. Since \texttt{PEUAF} has a unique shape, there might be concerns that such an oscillating function could be difficult to optimize. To address this, we compare the training dynamics of \texttt{PEUAF} and \texttt{ReLU}. Figure~\ref{fig:Loss of Baselines in three datasets} shows the training and validation curves of \texttt{PEUAF} and \texttt{ReLU} on the CWRU, PQD, and MF datasets. Below are our detailed analyses:

\begin{enumerate}
    \item \textbf{Convergence speed during training}: The convergence rate during the training process is notably influenced by the choice of activation functions and the inherent characteristics of the dataset. All the experiments in Figure~\ref{fig:Loss of Baselines in three datasets} consistently demonstrate the superior convergence speed of \texttt{PEUAF}. In dataset with a lower signal-to-noise ratio (such as the PQD dataset), \texttt{PEUAF} and \texttt{ReLU} show similar convergence speed. In contrast, in noise-free datasets or those with high signal-to-noise ratio, models adopting the \texttt{PEUAF} activation function display significantly faster convergence.

    \item \textbf{Convergence effect during training}: The choice of activation functions can impact the convergence effect, particularly in terms of oscillations or fluctuations during the training process. In the PQD dataset, the convergence patterns of \texttt{PEUAF} and \texttt{ReLU} are relatively similar, except for some fluctuations occurring around the 50th epoch. However, for the MF dataset, noticeable oscillations occur during convergence, particularly within the epoch range between 150 to 200. For the CWRU dataset, the fluctuations happen at around the 50th epoch when using \texttt{ReLU} as the activation function. Therefore, \texttt{PEUAF} helps reduce the oscillation of training losses and improves the training performance.

    \item \textbf{Fluctuation during validation}: In addition to the training process, the effectiveness of \texttt{PEUAF} can also be observed during the validation process. Across all the datasets, \texttt{PEUAF} outperforms \texttt{ReLU} by showing less fluctuation in the validation process. For the CWRU dataset, both \texttt{PEUAF} and \texttt{ReLU} exhibit fluctuations at the start of the validation process. However, after a sudden drop in validation loss after approximately 20 epochs, \texttt{PEUAF} shows smaller validation loss fluctuations than \texttt{ReLU}. For the PQD dataset, the validation loss curve for \texttt{PEUAF} and \texttt{ReLU} appear similar, but the amplitude of fluctuations is smaller for \texttt{PEUAF}. The most significant difference in fluctuation patterns is particularly obvious in the MF dataset, where \texttt{ReLU} exhibits high frequency and amplitude of fluctuations. This behavior can potentially be attributed to the fact that, in noise-free data settings, \texttt{ReLU} tends to capture global features initially, rather than precisely pinpointing fine-grained fault details, unlike \texttt{PEUAF}.
    
\end{enumerate}

% It is important to note that, despite the fluctuation observed during the training process with \texttt{ReLU}, the overall convergence rate remains favorable (MF). However, as the network architecture transitions to different typologies, the effectiveness of \texttt{PEUAF} gradually increases, ultimately resulting in superior convergence performance. Conversely, when the dataset exhibits a lower signal-to-noise ratio (PQD dataset), the convergence speed of both \texttt{PEUAF} and \texttt{ReLU} across the four network typologies demonstrates a relative parity. In stark contrast, the CWRU dataset consistently favors \texttt{PEUAF} over \texttt{ReLU} across all experimental. Particularly noteworthy is the Baseline (Figure~\ref{fig:Loss for CWRU Baseline FEUAF} and \ref{fig:Loss for CWRU Baseline ReLU}) and Inception (Figure~\ref{fig:Loss for CWRU VGG FEUAF} and \ref{fig:Loss for CWRU VGG ReLU}) network architectures, which display significantly accelerated convergence when adopting the \texttt{PEUAF} activation function. 

\subsection{Comparative Experiments }

In this section, we demonstrate combining the super-expressive activation function (\texttt{PEUAF}) with the baseline activation function can enhance the generalization ability of neural networks.  

\textbf{CIFAR-10.} In this experiment, we augment the dataset by rotating, shifting, shearing, and horizontally flipping the original images. We mainly focus on the ResNet structure. Table~\ref{Table: CIFAR10 eight structures} compares ResNet-18a with a mixed activation function to several identical model topologies using \texttt{ReLU}. The mixed activation function achieves a 0.89$\%$ error reduction. Tables~\ref{Table:CIFAR10} and ~\ref{Table:VitB16} separately summarize the test accuracy of ResNet-18 and Vit-B/16~\citep{dosovitskiy2020image} across various baseline activation functions and mixed activation functions. Notably, the mixed activation function improves the test accuracy, especially in \texttt{Softplus}, which increased by 2.72$\%$ and 5.01$\%$. 

\begin{table}[ht]
\centering\footnotesize
\caption{CIFAR-10 Classifcation error vs the number of parameters, for common compact model architectures vs.
ResNet-18a + Mixed ReLU.}
\label{Table: CIFAR10 eight structures}
\renewcommand{\arraystretch}{1.2}
\setlength{\tabcolsep}{5pt}
\begin{tabular}{l l l }
\toprule
Neural Network & $\#$Param & Error$\%$  \\
\midrule
All-CNN~\citep{Springenberg2014StrivingFS} & 1.3M & 7.25$\%$ \\
MobileNetV1~\citep{howard2017mobilenets} & 3.2M & 10.76$\%$ \\
MobileNetV2~\citep{sandler2018mobilenetv2} & 2.24M & 7.22$\%$  \\
ShuffleNet 8G~\citep{zhang2018shufflenet} & 0.91M  & 7.71$\%$  \\
ShuffleNet 1G~\citep{zhang2018shufflenet}  & 0.24M &  8.56$\%$   \\ 
HENet~\citep{duan2018speed} & 0.7M & 10.16$\%$   \\
ResNet-18a+ReLU & 0.27M & 8.75$\%$\\
ResNet-18a+ mixed ReLU &  0.27M &  7.82$\%$   \\
\bottomrule
\end{tabular}
\vspace{-0.3cm}
\end{table}

\begin{figure}[!t] 
\centering
\includegraphics[width=\linewidth]{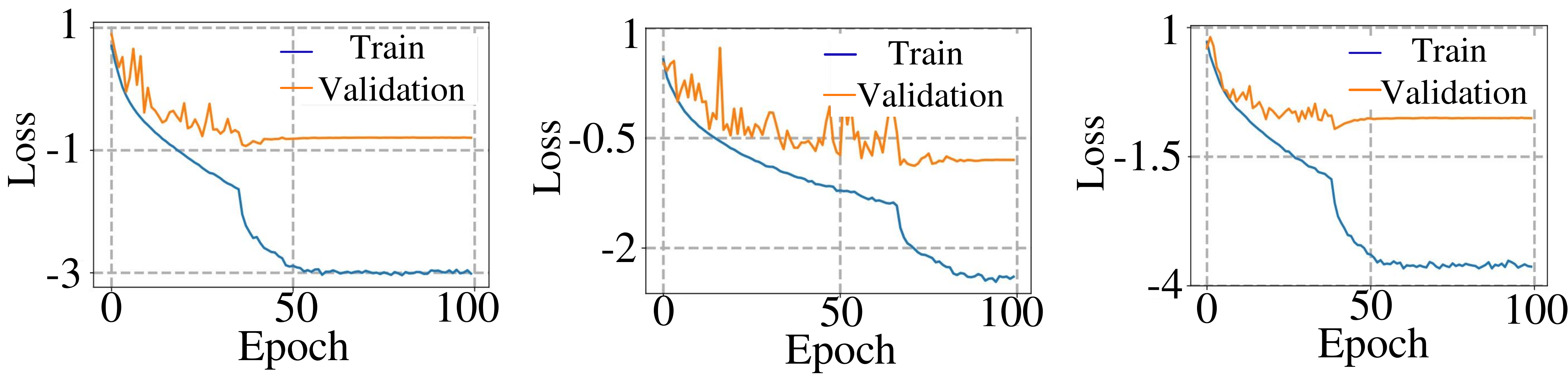}
% \subfloat[]{
% \includegraphics[width=0.3\linewidth]{figure/Loss for Resnet18 5Relu CIFAR10.pdf}
% \label{fig:Loss for Resnet18 5Relu}
% }~ \vspace{-0.1cm} \hspace{-2mm} 
% \subfloat[]{
% \includegraphics[width=0.31\linewidth]{figure/Loss for Resnet18 5MyActivation CIFAR10.pdf}
% \label{fig:Loss for Resnet18 5FEUAF}
% }~ \vspace{-0.1cm}   \hspace{-3.8mm} 
% \subfloat[]{
% \includegraphics[width=0.315\linewidth]{figure/Loss for Resnet18 4Relu+1Mayactivation CIFAR10.pdf}
% \label{fig:Loss for Resnet18 4Relu+1Mayactivation}
% }\vspace{-0.1cm} \\
% \subfloat[]{
% \includegraphics[width=0.3\linewidth]{figure/Accuracy for Resnet18 5Relu CIFAR10.pdf}
% \label{fig:Accuracy for Resnet18 5Relu}
% }~\vspace{-0.05cm} 
% \subfloat[]{
% \includegraphics[width=0.31\linewidth]{figure/Accuracy for Resnet18 5MyActivation CIFAR10.pdf}
% \label{fig:Accuracy for Resnet18 5FEUAF}
% }~\vspace{-0.05cm} 
% \subfloat[]{
% \includegraphics[width=0.3\linewidth]{figure/Accuracy for Resnet18 4Relu+1Mayactivation CIFAR10.pdf}
% \label{fig:Accuracy for Resnet18 4Relu+1Mayactivation}
% }\vspace{-0.05cm} \\
\caption{Loss of CIFAR-10 experiments among three activation functions. From left to right are the loss of ResNet-18 using \texttt{ReLU}, \texttt{PEUAF} and mixed activation function.}
\label{fig:Loss accuracy of 2D convolution experiments}
\vspace{-0.4cm}
\end{figure}

To further explain the efficacy of mixed activation functions, Figure~\ref{fig:Loss accuracy of 2D convolution experiments} provides a detailed comparison of the loss and accuracy among \texttt{ReLU}, \texttt{PEUAF} and mixed activation functions. When exclusively applying \texttt{PEUAF} in the CIFAR-10 classification task, both the training convergence and fluctuations are worse than those of \texttt{ReLU}, as shown in the loss curve in Figure~\ref{fig:Loss accuracy of 2D convolution experiments}. However, the ResNet-18 using mixed activation functions outperforms the models using either \texttt{ReLU} or \texttt{PEUAF} alone. The mixed approach results in smoother loss and accuracy curves during both the training and validation process. 

\begin{figure}[ht] 
\flushleft
\subfloat[]{
\includegraphics[width=0.45\linewidth]{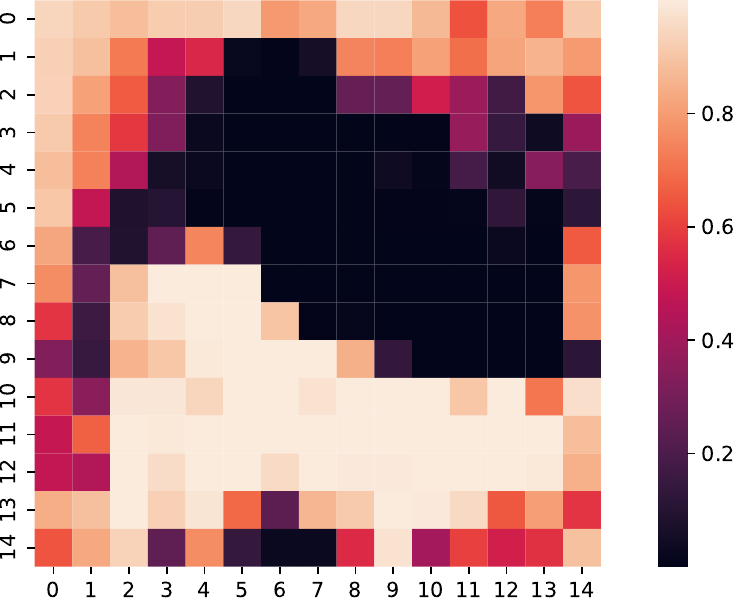}
\label{fig:Occlusion_exp for Resnet18 5Relu}
}~\vspace{-0.2cm}
\subfloat[]{
\includegraphics[width=0.45\linewidth]{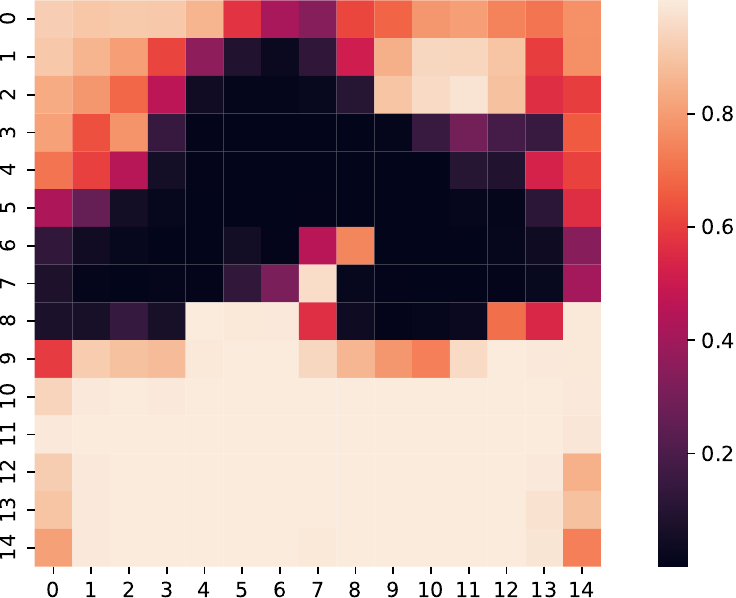}
\label{fig:Occlusion_exp for Resnet18 5FEUAF}
}\\
\subfloat[]{
\includegraphics[width=0.45\linewidth]{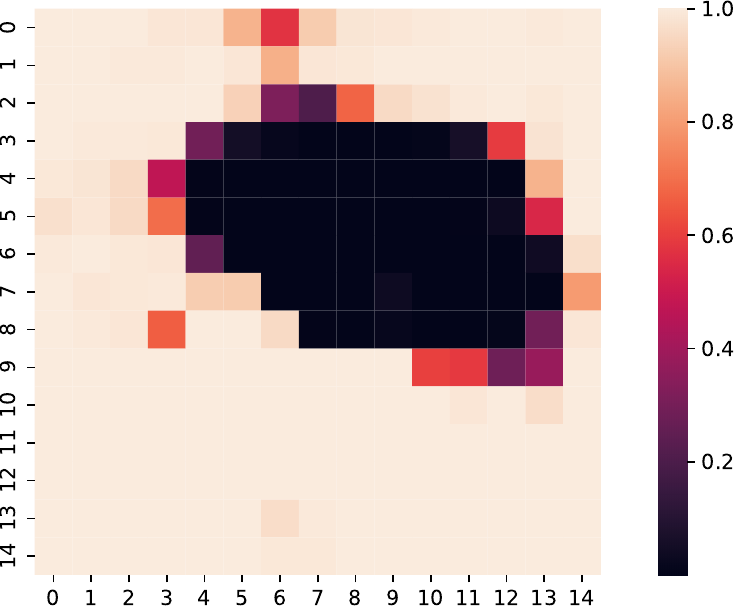}
\label{fig:Occlusion_exp for Resnet18 4Relu+1Mayactivation}
}~\vspace{-0.05cm}
\subfloat[]{
\includegraphics[width=0.4\linewidth]{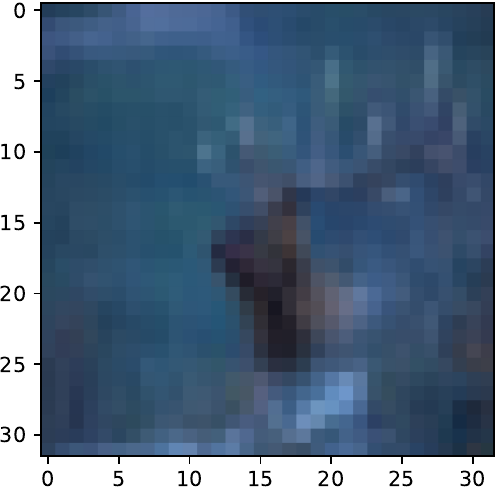}
\label{fig:Original pic for Resnet18 4Relu+1Mayactivation}
}\vspace{-0.05cm}\\
\caption{Occlusion experiments on the CIFAR-10 dataset among three activation functions to examine their discriminative ability. (a)$\sim$(c) Occlusion experiments using \texttt{ReLU}, \texttt{PEUAF} and mixed activation function. (d) the original figure.}
\label{fig:Occlusion experiments of 2D convolution experiments}
\vspace{-0.4cm}
\end{figure}

The occlusion experiments further demonstrate that the mixed activation functions can enhance the neural network's ability to identify essential features. The occlusion sizes and strides are set to 4 and 2, respectively, with occluded pixels replaced by zeros. As in Figure~\ref{fig:Occlusion experiments of 2D convolution experiments}, the results provide a clear illustration of this phenomenon. The models using only \texttt{ReLU} or \texttt{PEUAF} successfully identify a multitude of features contributing to the classification. However, they also select too many unnecessary pixel points, recognizing part of the surroundings as the important features for classification. In contrast, the mixed activation function model can accurately locate the critical features while ignoring irrelevant pixels.

% The parameters such as occluding size and stride are chosen based on the length of signals in datasets.

\begin{table}[htbp]
\centering\footnotesize
\caption{Comparisons of classification accuracy across several activation functions using ResNet for CIFAR-10.}
\label{Table:CIFAR10}
\renewcommand{\arraystretch}{1.2}
\setlength{\tabcolsep}{5pt}
\begin{tabular}{l l }
\toprule
Activation & Test accuracy  \\
\midrule
ResNet-18+PEUAF & 90.00$\%$ / - \\
ResNet-18+LReLU/Mixed & 92.42$\%$ / 94.13$\%$ \\
ResNet-18+PReLU/Mixed & 92.29$\%$ / 94.23$\%$  \\
ResNet-18+Softplus/Mixed  & 89.28$\%$ / 92.09$\%$  \\
ResNet-18+ELU/Mixed  & 91.09$\%$ / 92.11$\%$   \\
ResNet-18+SELU/Mixed  & 90.47$\%$ / 91.32$\%$   \\
ResNet-18+ReLU/Mixed  & 93.02$\%$ / 93.91$\%$  \\
ResNet-18+Swish/Mixed  & 94.07$\%$ / 92.99$\%$   \\
ResNet-34+ReLU/Mixed  & 93.70$\%$ / 94.23$\%$   \\
% \makecell[l]{Vit-transformer pruning\\+GELU/Mixed \citep{zhu2021vision}} &80.00$\%$ / 81.52$\%$\\
\bottomrule
\end{tabular}
\vspace{-0.3cm}
\end{table}

\begin{table}[htbp]
\centering\footnotesize
\caption{Comparisons of classification accuracy across several activation functions using Vit-B/16 for CIFAR-10.}
\label{Table:VitB16}
\renewcommand{\arraystretch}{1.2}
\setlength{\tabcolsep}{5pt}
\begin{tabular}{l l }
\toprule
Activation & Test accuracy  \\
\midrule
Vit-B/16+PEUAF & 90.58$\%$ / - \\
Vit-B/16+LReLU/Mixed & 91.15$\%$ / 91.31$\%$ \\
Vit-B/16+PReLU/Mixed & 90.40$\%$ / 90.47$\%$  \\
Vit-B/16+Softplus/Mixed  & 74.23$\%$ / 79.24$\%$  \\
Vit-B/16+ELU/Mixed  & 89.69$\%$ / 89.80$\%$   \\
Vit-B/16+SELU/Mixed  & 87.26$\%$ / 87.37$\%$   \\
Vit-B/16+ReLU/Mixed  & 89.43$\%$ / 89.44$\%$  \\
Vit-B/16+Swish/Mixed  & 90.66$\%$ / 89.65$\%$   \\
Vit-B/16+GELU/Mixed  & 97.49$\%$ / 97.90$\%$   \\
\bottomrule
\end{tabular}
\vspace{-0.3cm}
\end{table}

\textbf{Tiny-ImageNet}. The Tiny-Imagenet dataset is utilized to further demonstrate the expressiveness of \texttt{PEUAF}. The model is trained for 100 epochs with an initial learning rate of 0.1, which decays by an order of magnitude every 30 epochs, using a batch size of 256. Table~\ref{Table:TinyImageNet} compares the test accuracy of ResNet-18 with several baseline activation functions on Tiny-ImageNet. By replacing the activation functions in the last block, the ResNet-18 with mixed activation functions achieves competitive results, showing slight improvements in the test accuracy across most experiments, except for \texttt{Swish} and \texttt{PEUAF}.

\begin{table}[!t]
\centering\footnotesize
\caption{Comparisons of classification accuracy across several activation functions using ResNet-18 for Tiny-ImageNet.}
\label{Table:TinyImageNet}
\renewcommand{\arraystretch}{1.2}
\setlength{\tabcolsep}{5pt}
\begin{tabular}{l l }
\toprule
Activation & Test accuracy  \\
\midrule
ResNet-18+PEUAF & 56.86$\%$ / - \\
ResNet-18+LReLU/Mixed & 62.39$\%$ / 62.29$\%$ \\
ResNet-18+PReLU/Mixed & 59.57$\%$ / 60.81$\%$  \\
ResNet-18+Softplus/Mixed  & 56.98$\%$ / 57.75$\%$  \\
ResNet-18+ELU/Mixed  & 59.44$\%$ / 59.88$\%$   \\
ResNet-18+SELU/Mixed  & 59.51$\%$ / 59.62$\%$   \\
ResNet-18+ReLU/Mixed  & 63.40$\%$ / 63.42$\%$  \\
ResNet-18+Swish/Mixed  & 60.76$\%$ / 59.53$\%$   \\
\bottomrule
\end{tabular}
\vspace{-0.3cm}
\end{table}

\textbf{ImageNet}. The ImageNet dataset is used to evaluate the effectiveness of \texttt{PEUAF} in large datasets. Due to memory limitations, the model follows the setup from the previous research~\citep{liu2022automix}, except for the number of neurons in the first layer and the data enhancement. The neurons of the first layer is reduced to 256 from 512. Table~\ref{Table:ImageNet} compares the test accuracy between \texttt{PReLU} and the mixed activation. In this large-scale image classification experiment, the drawbacks of using \texttt{PEUAF} alone become apparent, with the accuracy of ResNet-18 with \texttt{PEUAF} being only 63.38\%, which is lower than that of \texttt{PReLU}. However, the ResNet-18 with mixed activation functions achieves competitive results. 

\begin{table}[htbp]
\centering\footnotesize
\caption{Comparisons of classification accuracy across several activation functions using ResNet-18 for ImageNet.}
\label{Table:ImageNet}
\renewcommand{\arraystretch}{1.2}
\setlength{\tabcolsep}{5pt}
\begin{tabular}{l l }
\toprule
Activation & Test accuracy  \\
\midrule
ResNet-18+PEUAF & 63.38$\%$ / - \\
% ResNet-18+ReLU/Mixed & 71.28$\%$ / 70.32$\%$ \\
ResNet-18+LReLU/Mixed & 70.65$\%$ / 70.96$\%$ \\
% ResNet-18+Swish/Mixed & 71.23$\%$ / 71.02$\%$ \\
\bottomrule
\end{tabular}
\vspace{-0.3cm}
\end{table}

\section{Conclusion and Discussion}

This paper provides an in-depth analysis of the characteristics and effectiveness of \texttt{PEUAF}, particularly focusing on its application to industrial and image datasets. By testing the trainable frequency \( w \), we have determined an optimal frequency range for \( w \) within the interval \([0, 1]\). To further demonstrate the super-expressiveness of \texttt{PEUAF}, we have conducted experiments using four industrial datasets and three benchmark image datasets. The results indicate that \texttt{PEUAF} surpasses \texttt{ReLU} in terms of convergence speed, oscillation during training, fluctuation during validation, and fault localization ability, especially in industrial datasets with a high signal-to-noise ratio. Additionally, the mixed activation function outperforms the single activation function in most image classification tasks.

Looking ahead, the future of activation function research is promising. The development of \texttt{PEUAF} paves the way for exploring other super-expressive activation functions that could further enhance neural network performance across various applications. Future research could focus on expanding the family of super-expressive activation functions and investigating their practical utility in more diverse and complex datasets. Moreover, combining \texttt{PEUAF} with other state-of-the-art neural network architectures and exploring its benefits in real-world scenarios could yield valuable insights. The adaptability and effectiveness of \texttt{PEUAF} in handling stationary signals suggest potential applications in fields such as signal processing, fault diagnosis, and time-series analysis.

% By testing the trainable frequency $w$ and diverse activation functions in negative side, we have identified an optimal frequency range for $w$, falling within the interval $[0, 1]$. Furthermore, the ${softsign}$ activation function in the negative side outperforms \texttt{ReLU}, \texttt{Swish} and ${Tanh}$. In addition to the characteristics investigations, three kinds of industrial field datasets, and neural network models with four topological structures are utilized to further substantiate the super-expressiveness of \texttt{PEUAF} within fault diagnosis tasks. The converge speed and oscillation in training process, the fluctuation in validation process, and the ability of fault locating in \texttt{PEUAF} all outperforms \texttt{ReLU} particularly in dataset with high signal-noise ratio. Moreover, the mixed activation function outperforms the single activation function in CIFAR10 classification. Importantly, only limited neural units with super-expressive activation function are required to help the neural network locate the essential features while minimizing the inclusion of extraneous or irrelevant elements.

% \bibliographystyle{elsarticle-harv}
% % \bibliographystyle{natbib}
% % \biboptions{authoryear}
% % \bibliographystyle{ieeetr}
% \bibliography{reference}

\end{document}